\pdfoutput=1

\documentclass[11pt]{article}

\usepackage[final]{acl}

\usepackage{times}
\usepackage{latexsym}

\usepackage{amsmath,amsfonts,amssymb}
\usepackage{tabularx}
\usepackage{pdfpages}
\usepackage{amssymb}
\usepackage{amsmath}
\usepackage{bm}
\usepackage{graphicx}
\usepackage{booktabs}
\usepackage{mathtools}
\usepackage{multirow}
\usepackage{multicol}
\usepackage{makecell}
\usepackage{diagbox}
\usepackage{enumitem}
\usepackage{array}
\usepackage{afterpage}
\usepackage{float}

\usepackage{colortbl}      
\usepackage{array}

\usepackage{framed}
\usepackage{mdframed}
\usepackage{tcolorbox}
\usepackage{multicol}
\usepackage{xcolor} 
\definecolor{myblue}{HTML}{1020BB} 
\definecolor{mywhite}{HTML}{FFFFFF}

\usepackage[T1]{fontenc}

\usepackage[utf8]{inputenc}

\usepackage{microtype}

\usepackage{inconsolata}

\usepackage{graphicx}

\title{AirRAG: Autonomous Strategic Planning and Reasoning Steer Retrieval Augmented Generation}

\author{Wenfeng Feng\thanks{The first two authors contributed equally}, Chuzhan Hao\footnotemark[1], Yuewei Zhang, Guochao Jiang, Jingyi Song \& Hao Wang\protect \thanks{ Corresponding author}
\\
Alibaba Cloud Computing
\\
\tt {\{wenfeng.fwf,haochuzhan.hcz\}@alibaba-inc.com,cashenry@126.com}
}

\begin{document}
\maketitle
\begin{abstract}
Leveraging the autonomous decision-making capabilities of large language models (LLMs) has demonstrated superior performance in reasoning tasks. However, despite the success of iterative or agentic retrieval-augmented generation (RAG) techniques, these methods are often constrained to a single solution space when confronted with complex problems.
In this paper, we propose a novel thinking pattern in RAG that integrates autonomous strategic planning with efficient reasoning actions, significantly activating intrinsic reasoning capabilities and expanding the solution space of specific tasks via Monte Carlo Tree Search (MCTS), which we refer to as \textbf{AirRAG}. Specifically, our approach designs five fundamental reasoning actions, which are expanded to a broad tree-based reasoning space using MCTS.
The approach also incorporates self-consistency verification to explore potential reasoning paths and inference scaling law. Additionally, computationally optimal strategies are employed to allocate more inference resources to key actions, thereby enhancing overall performance. Experimental results demonstrate the effectiveness of AirRAG, showing significant performance gains on complex question-answering datasets. Furthermore, AirRAG is flexible and lightweight, making it easy to integrate with other advanced technologies and models.

\end{abstract}

\begin{figure}[htbp]
\centering
\includegraphics[width=0.48\textwidth]{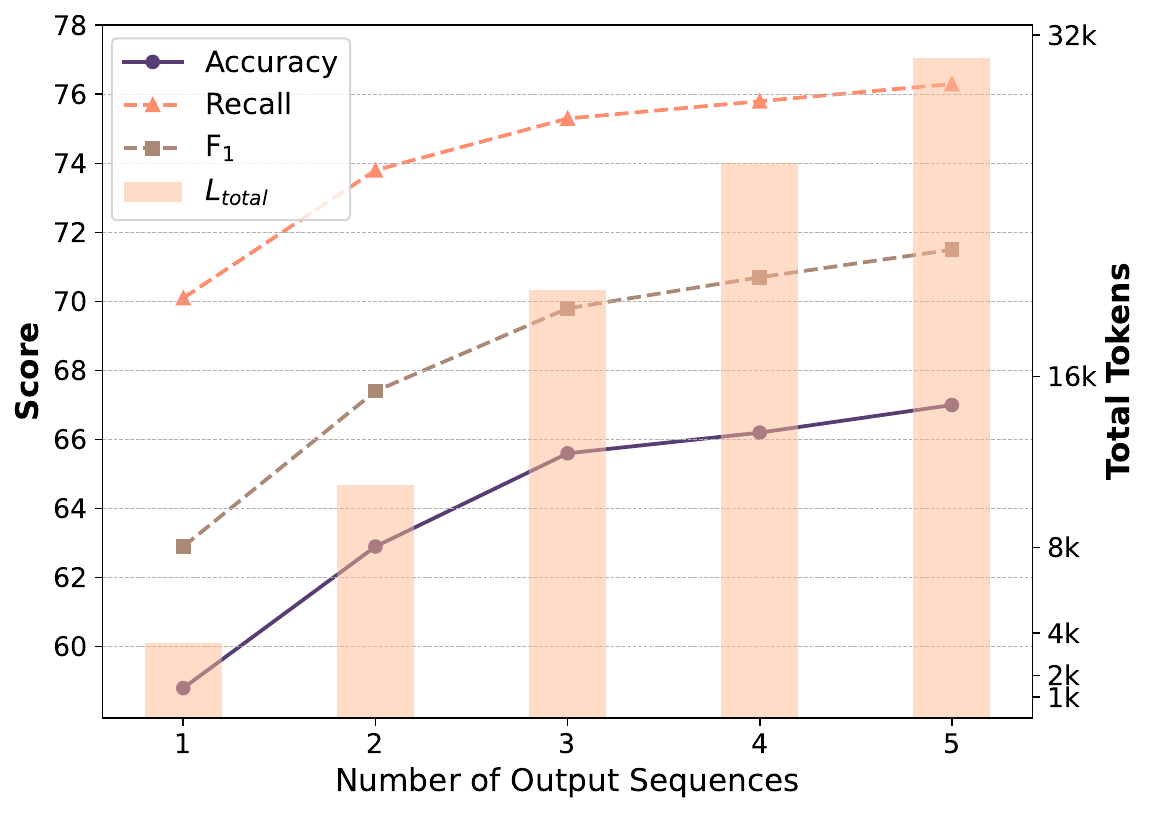}
\caption{Comparison of average performance across three datasets with varying numbers of output sequences. $L_{\text{total}}$ represents the total number of tokens consumed during the reasoning process. AirRAG leverages generation diversity and self-consistency to explore the potential solution space, significantly enhancing overall performance by scaling inference computation.}
\label{fig:intro}
\end{figure}
\section{Introduction}
Retrieval-Augmented Generation (RAG) has shown great potential in addressing the issue of generating factually incorrect content, especially in domain-specific or knowledge-intensive tasks~\citep{pmlr-v202-kandpal23a}. However, as task complexity increases, several new challenges emerge, such as the inability to retrieve sufficient knowledge with a single query and the difficulty of understanding the intricate reasoning logic inherent in the question.
To tackle these challenges, it is crucial to harness the reasoning capabilities of large language models (LLMs) to improve RAG performance~\citep{jiang-etal-2023-active,jeong-etal-2024-adaptive,asai-2024-selfrag,yu2024autorag}.

Previous research on complex query scenarios has primarily focused on optimizing the query and retrieval processes to obtain relevant information~\citep{pmlr-v202-shi23a,zhou2023leasttomost,gao-etal-2023-precise-hyde,jiang-etal-2023-active,zheng2024takeastepback,asai-2024-selfrag,yan2024corrective}. Iterative retrieval is frequently used to improve the depth and relevance of search results in information retrieval tasks. This process continuously updates intermediate queries and results to satisfy dynamic information needs during the complex task-solving process~\citep{jeong-etal-2024-adaptive,yue2024iterdrag}. In addition, \citet{li2025searcho1} integrates an
agentic search workflow into the reasoning process, enabling dynamic retrieval when LLMs encounter uncertain knowledge points. The agentic LLMs are trained to learn step-by-step reasoning with search through reinforcement learning~\citep{chen2025research,zheng2025deepresearcher}.

However, these approaches face two significant issues. First, the single reasoning paradigm and the chain-like reasoning process often fail to effectively explore the solution space, particularly when reasoning relies on self-exploration. This process is vulnerable to low-quality intermediate reasoning steps and is easily trapped in a narrow solution space. Second, the agentic search workflow and guiding self-exploration become challenging when using relatively smaller language models (e.g., \textit{Qwen2.5-7B-Instruct}~\citep{qwen2.5}). Furthermore, trainable agentic LLMs require efficient reinforcement learning training data and are difficult to apply to models with hundreds of billions of parameters.

In response to these challenges, we propose AirRAG, a method that leverages autonomous strategic planning and reasoning capabilities and expands the solution space using Monte Carlo Tree Search (MCTS). 
We design five fundamental reasoning actions: system analysis, direct answer, retrieval-answer, query transformation, and summary-answer. These actions are the core and frequently used ones in the deep search scenarios, which can effectively address a wide range of problems in various scenarios, including those that require progressive or parallel queries. Importantly, these actions can be executed efficiently on LLMs of different scales. Additionally, we introduce MCTS and self-consistency to enable controllable reasoning path generation and efficient inference scaling. To accurately select the answer from multiple reasoning paths, we combine a voting mechanism with a process-supervised reward model. As inference computation increases, our approach demonstrates significant performance improvements as shown in Figure~\ref{fig:intro}. Moreover, AirRAG features a flexible architecture that can easily integrate other advanced methods into the approach as additional action branches.
In summary, our main contributions are as follows:
\begin{itemize}
\item We design five fundamental reasoning actions that can address most problem types in deep search scenarios, ensuring controllable reasoning processes.

\item We introduce MCTS and self-consistency to effectively expand the solution space for complex tasks. 
Our approach improves generalization and performance through comprehensive inference scaling and a pluggable architecture.

\item
We show thorough experimental results that AirRAG outperforms current iterative or agentic methods, effectively activating the planning and reasoning capabilities of LLMs and flexibly expanding the solution space.
\end{itemize}

\section{Related Work}
\textbf{Retrieval-Augmented Generation (RAG)}. RAG has demonstrated significant improvements in the performance of LLMs in knowledge-intensive tasks. Compared to vanilla RAG, optimizing the query and retrieval process enhances knowledge correlation and, consequently, improves reasoning performance. Several methods, such as query expansion and transformation, have been proposed to achieve better retrieval results~\citep{zhou2023leasttomost,ma-etal-2023-query,gao-etal-2023-precise-hyde}. However, as task complexity increases, retrieving sufficient knowledge in a single query becomes increasingly difficult. To address this, iterative retrieval techniques have been proposed to gather additional contextual references. For instance, IRCoT~\citep{trivedi-etal-2023-ircot} utilizes chain-of-thought (CoT) to guide the retrieval process, refining the CoT with the retrieved information. Similarly, ITER-RETGEN~\citep{shao-etal-2023-iterretgen} combines retrieval and generation modules to promote a deeper understanding of specific tasks.

\noindent\textbf{Autonomous Planning and Reasoning in RAG}. In addition to optimizing retrieval, activating the planning and reasoning capabilities of LLMs can significantly improve the efficiency and relevance of the retrieved information. 
Leveraging the decision-making abilities of LLMs enhances the overall performance~\citep{nakano-2022webgpt,schick-2023-toolformer}. 
Self-RAG and its variants~\citep{asai-2024-selfrag,yan2024corrective,jeong-etal-2024-adaptive} adopt a self-reflection mechanism that iteratively predicts reflection tokens during training, enabling better control during inference. Auto-RAG~\citep{yu2024autorag} systematically plans retrievals and refines queries to acquire valuable knowledge through multi-turn iterations. IterDRAG~\citep{yue2024iterdrag} explores inference scaling strategies in RAG, improving LLMs’ ability to effectively acquire and utilize contextual information. Search-o1~\citep{li2025searcho1} designs an agentic search workflow to dynamically obtain effective knowledge. ReSearch~\citep{chen2025research} and DeepResearcher~\citep{zheng2025deepresearcher} train agentic LLMs to reason with search using reinforcement learning. Despite the progress made in these methods, they often struggle to explore the solution space effectively during reasoning. Self-exploration frequently leads to being trapped in a limited solution space, hindered by low-quality reasoning steps even after multiple iterations. This issue is often attributed to the chain reasoning pattern and the difficulty small-scale LLMs face when handling overly complex tasks in a single iteration.

\begin{figure*}[ht]
\centering
\includegraphics[width=0.99\textwidth]{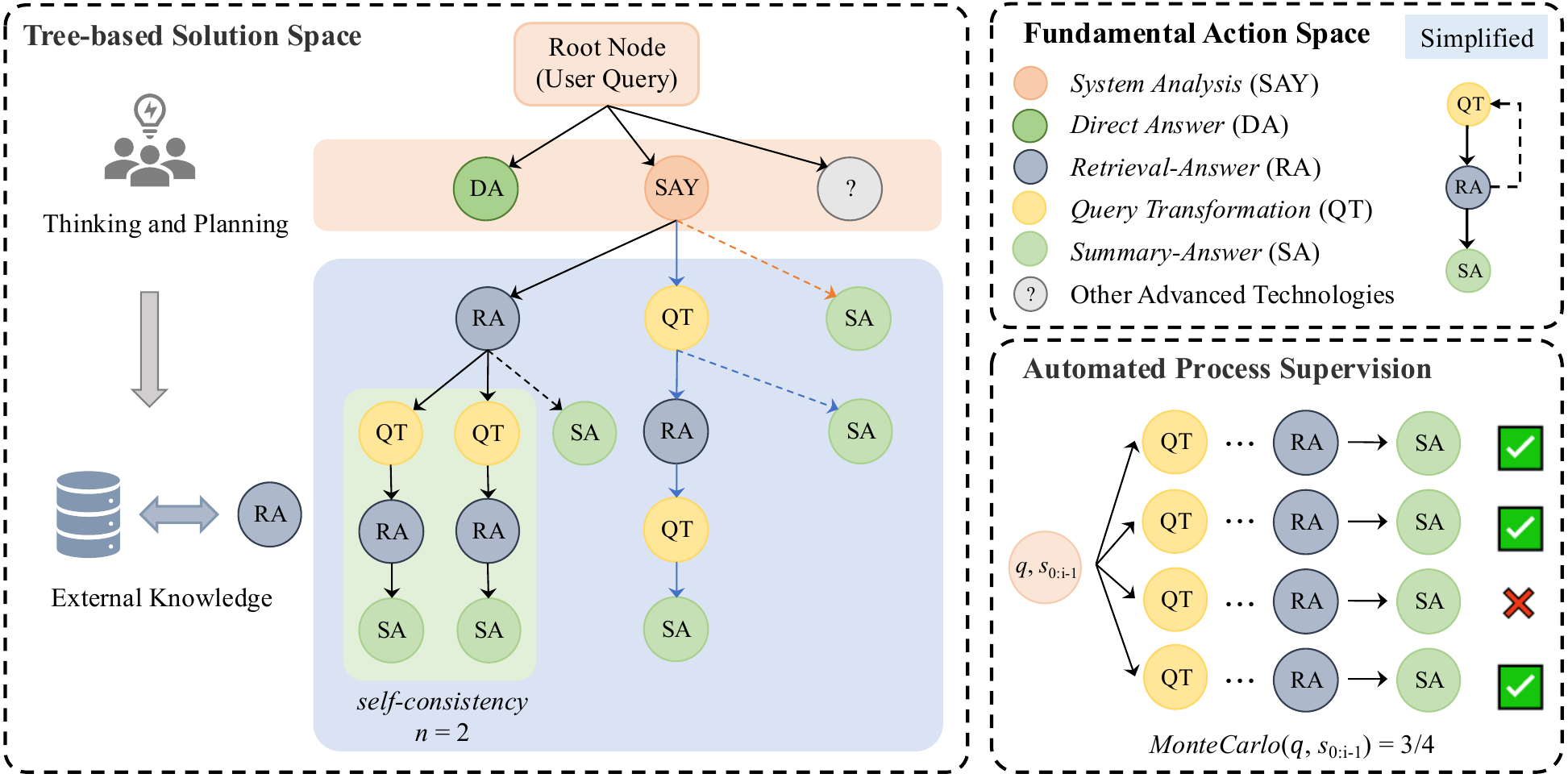}
\caption{The schematic diagram of our proposed AirRAG. AirRAG implements a paradigm that combines system thinking with step-by-step reasoning. In the inference phase, we introduce MCTS and self-consistency to scaling computation, which significantly outperforms other strong baselines.}
\label{fig:framework}
\end{figure*}

\noindent\textbf{Monte Carlo Tree Search (MCTS)}. To address these challenges, tree-based search algorithms, particularly Monte Carlo Tree Search (MCTS), have emerged as effective tools to expand search spaces and enhance reasoning capabilities~\citep{silver2017mastering,chen2024alphamath,qi2024rstar,zhang2024llamaberry}. MCTS has been shown to extend reasoning by exploring multiple branching queries, thus enabling the exploration of diverse reasoning paths~\citep{yao2023ToT,besta2024got}. In the mathematical reasoning scenario, \citet{zhang2024llamaberry} and \citet{chen2024alphamath} leverage MCTS to achieve more efficient exploration of solution spaces, while \citet{qi2024rstar} designs rich human-like reasoning actions to improve reasoning trajectories. Furthermore, recent research indicates that inference scaling~\citep{yue2024iterdrag} and self-consistency~\citep{wang2023selfconsistency} can lead to substantial improvements. In this context, our approach samples diverse reasoning paths to achieve both inference scaling and self-consistency verification during the next expansion of the action space.

Unlike existing methods that focus on optimizing query and retrieval processes or leveraging LLMs' reasoning capabilities through iterative retrieval, AirRAG uniquely integrates MCTS and self-consistency to systematically expand the solution space and ensure the controllability of the reasoning process. Simultaneously, we design five fundamental reasoning actions that effectively address a broader range of question types, particularly in complex tasks. 
This pluggable architecture also allows for easy integration of current advanced methods or reasoning language models, making AirRAG a flexible and powerful solution for deep search scenarios.
In the experiment, we thoroughly verify the performance gains brought by the inference scaling law and investigate how to rationally allocate inference resources.

\section{Methodology}
In order to effectively explore the solution space during reasoning, we propose a controllable tree-based framework of RAG. This framework combines Monte Carlo Tree Search (MCTS) with five distinct reasoning actions, enabling efficient and controlled expansion of the solution space. Meanwhile, we further implement more comprehensive inference scaling strategies based on~\citet{yue2024iterdrag} and employ pruning techniques along with computationally optimal strategies to strike a balance between effectiveness and efficiency. The whole process is illustrated in Figure~\ref{fig:framework}.

\subsection{Define Fundamental Reasoning Actions}\label{sec:fundamental-actions}
Relying solely on the autonomy of LLMs for iterative self-exploration often results in getting trapped in a solution space that is difficult to navigate, especially when dealing with different types of complex questions. IterDRAG~\citep{yue2024iterdrag} uses a single action type to generate the next reasoning step, which can lead to ineffective space exploration. The core of MCTS generation lies in the action space, which defines the scope of tree exploration. 
Based on advanced methods and reasoning language models, we summarize the most common actions in RAG, such as query transformation and retrieval answering. Meanwhile, the chain-of-thought in reasoning models has demonstrated superior performance in complex open-domain question answering. 
Therefore, simplifying human cognitive processes in complex reasoning is essential~\citep{jaffe2023modelling}. Inspired by this, we introduce five fundamental human-like reasoning actions to bridge the gap between LLM reasoning and human cognition in RAG scenarios.

\begin{itemize}[leftmargin=*,itemindent=0pt]
    \item \noindent\textit{$A_1$}: \textit{System Analysis} (SAY). This action analyzes the overall structure of the problem, followed by its decomposition or planning. It represents systematic and global thinking before problem-solving.
    \item \noindent\textit{$A_2$}: \textit{Direct Answer} (DA). This action leverages parametric knowledge of LLMs to answer questions directly, without relying on any external knowledge.
    \item \noindent\textit{$A_3$}: \textit{Retrieval-Answer} (RA). This action retrieves related knowledge from the external knowledge base to support subsequent reasoning.
    \item \noindent\textit{$A_4$}: \textit{Query Transformation} (QT). This action transforms human questions in order to improve retrieval performance. It supports various transformations, such as rewriting, step back prompting, follow-up questions and multi-query retrieval.
    \item \noindent\textit{$A_5$}: \textit{Summary-Answer} (SA). This action combines intermediate reasoning steps, answers and the initial questions to generate the final answer.
\end{itemize}


The above five actions define a highly diverse action space $\{A_1, A_2, A_3, A_4, A_5\}$. In the first step, the initial state is denoted as $s_0$ and then MCTS selects the action $a_1$ and $a_2$ to prompt the LLM to generate the next reasoning steps in parallel. Subsequent actions are performed sequentially to expand the reasoning path. It is important to note that there are sequential dependencies between different actions. For example, $A_1$ and $A_2$ can only be executed after the root question. Additionally, we incorporate the diverse sampling of self-consistency~\citep{wang2023selfconsistency} for each action to expand the reasoning paths. Specifically, an action is more likely to generate the correct reasoning step if we sample multiple times in the current state. Finally, we can obtain multiple generated reasoning trajectories, such as $[s_0\oplus s_{1:n}]$. To further improve inference efficiency, we choose the action $\{A_3, A_4, A_5\}$ as a simplified action space, referred to as \textit{AirRAG-Lite}, which achieves a better balance between efficiency and effectiveness.

\subsection{Perform Reasoning Processes via MCTS}
\subsubsection{Solution Generation} 
Based on the action space defined above, we introduce MCTS to generate candidate reasoning trajectories. The initial root node, $s_0$, represents the question without any reasoning steps. The policy is directly modeled by a language model as $\pi(a|s)=\text{LM}(a|s)$, and the state transition function combines preceding reasoning steps with current actions, i.e., $s_{i}=\text{Concat}(s_{0:i-1}, a_{j})$. During each MCTS rollout, we execute multiple steps, including \textit{selection}, \textit{expansion}, \textit{simulations}, and \textit{backpropagation}. Multiple rollouts are performed to expand the solution space. To balance the exploration and exploitation, we adopt the well-known Upper Confidence Bounds applied to Trees (UCT)~\citep{UCT2006} for node selection as follows:
\begin{align}
\centering
    \text{UCT}(s,p) = \frac{Q(s,a)}{N(s)}+w\sqrt{\frac{\log N_p(s)}{N(s)}}, 
\end{align}
where $Q(s,a)$ is the reward value for node $s$, generated by taking action $a$, and is updated through backpropagation. $Q(s,a)$ serves as a key metric in MCTS to evaluate the value of each node. $N(s)$ denotes the number of visits to $s$, $p$ is the parent node of $s$, and $w$ is the weight to balance exploration and exploitation. Initially, all unexplored nodes are assigned $Q(s_i, a_i) = 0$, leading to random tree expansions at the beginning. When the search reaches a terminal node $n_d$, a reward score $Q(s_d, a_d)$ is computed based on whether the node reaches the correct answer. This score is then back-propagated to all intermediate nodes along the trajectory $t = x \oplus s_1 \oplus s_2 \oplus \ldots \oplus s_d$. Specifically, for each intermediate node $s_i$ (for $i = 1, 2, \ldots, d-1$), its $Q$ value is updated as: $Q(s_i, a_i) = Q(s_i, a_i) + Q(s_d, a_d)$.

When the search reaches a terminal node, defined either by a terminal state or a predetermined maximum tree depth $d$, we obtain a trajectory from the root to the terminal node. All trajectories from the rollout iterations are collected as candidate solutions. Section~\ref{select_verify} explains how we select the optimal answer node from these trajectories.

\subsubsection{Inference Scaling}\label{sec:3.2.2}
Numerous studies have demonstrated that scaling inference computation can significantly improve the performance of LLMs without additional training~\citep{snell2024scaling, yue2024iterdrag}. Based on the above methods, we explore strategies to leverage inference computation scaling in AirRAG. One straightforward strategy is extending the \textit{effective context length} (short for $L_{\max}$) during the document retrieval phase, allowing more related documents to supplement the knowledge base. Additionally, we perform multiple \textit{rollouts} to thoroughly explore the solution space relying on the tree-based search. Adjusting \textit{the number of output sequences} ($n$) generated during certain actions enables self-consistency verification and further inference scaling. These strategies provide flexibility for scaling inference computation in RAG, empowering LLMs to address complex knowledge-intensive queries more effectively.

To improve efficiency and minimize redundant computations, we implement an early pruning strategy for state nodes and reasoning paths. Deduplication is applied to the output sequence states generated by each action, ensuring the diversity of the subsequent path. Furthermore, if multiple rollouts select the same state sequence, only one valid reasoning path is retained.

\subsubsection{Flexible Architecture}
Our tree-based architecture provides the flexibility to integrate other advanced approaches. We reproduce the IterDRAG method based on the prompt design by~\citet{yue2024iterdrag}. Meanwhile, inspired by its iterative implementation, we simplify the fundamental action space to $\{A_3, A_4, A_5\}$, enabling a faster implementation while still achieving relatively good results. These methods serve as an exploratory extension of our approach and can be activated or deactivated as needed. Due to the training-free nature of our method, its generator LLM can be arbitrarily replaced with the strongest existing models for performance improvement.

\subsection{Select the Optimal Answer Node}\label{select_verify}
For common mathematical reasoning tasks, a simple consistency-based method can efficiently select the most precise reasoning path. For example, the most frequent number extracted from multiple candidate solutions in MATH~\citep{hendrycks2021MATH} can be chosen as the final answer. However, extracting precise answers and performing effective aggregation becomes more challenging for knowledge-intensive tasks. To address this, we design two self-consistency verification methods for such problems. \textit{Jaccard similarity} and \textit{text embeddings} are two different approaches used in natural language processing to measure the similarity between texts. We apply these methods to cluster text answers and compute answer scores as follows:
\begin{align}
\centering
    \text{jcdScore}_i&=\frac{1}{N}\sum_{j=1}^N \frac{| A_i \cap A_j |}{| A_i \cup A_j |}, \\
    \text{embScore}_i&=\frac{1}{N}\sum_{j=1}^N \cos(E_i, E_j),
\end{align}
where \textit{N} is the number of valid answer nodes, $A_i$ is the word-level set of answer text $i$, and $E_i$ denotes the embedding vector of answer text $i$.

In addition, we further investigate the \textit{self-refine} and process-supervision \textit{reward model} to identify the most accurate reasoning trajectory. Self-refinement uses the $A_5$ (Summary-Answer) action to refine the final answer from all candidate answer nodes. The reward modeling process consists of two steps: data synthesis and instruction tuning.
\begin{itemize}
    \item \textbf{Data synthesis}: We leverage MCTS to perform multiple rollouts on partial training sets. Based on known ground truth, we sample positive and negative reasoning trajectories and use Monte Carlo estimation to evaluate intermediate state scores.
    \item \textbf{Instruction tuning}: Synthetic samples are used to fine-tune a relatively small LLM, such as \textit{Qwen2.5-14B-Instruct}.
\end{itemize}

\section{Experiments}\label{sec:experiments}
In this section, we conducted experiments on complex QA benchmarks by answering the following research questions.
\begin{itemize}
\item \textbf{RQ1}: Does AirRAG outperform state-of-the-art baselines?

\item \textbf{RQ2}: How does AirRAG perform when it comes to comprehensive inference scaling?

\item \textbf{RQ3}:  What is the performance benefit of AirRAG in optimizing the allocation of inference computation?

\item \textbf{RQ4}: How does AirRAG perform for various verification methods for multiple candidate rollouts?

\item \textbf{RQ5}: What is the intuitive efficiency and performance of AirRAG in the reasoning process?
\end{itemize}
\subsection{Experimental Settings}
\subsubsection{Datasets}
To evaluate the effectiveness of AirRAG, we conduct experiments on various question-answering (QA) tasks, including both open-domain QA and multi-hop QA. 
The complex multi-hop QA datasets consist of HotpotQA~\citep{yang-etal-2018-hotpotqa}, MuSiQue~\cite{trivedi-etal-2022-MuSiQue} and 2WikiMultiHopQA (2Wiki) ~\citep{ho-etal-2020-2wiki}.
Other single-hop QA datasets include Natural Questions (NQ)~\citep{kwiatkowski-etal-2019-naturalQA}, TriviaQA~\citep{joshi-etal-2017-triviaqa}, PopQA~\citep{mallen-etal-2023-popqa} and WebQA~\citep{berant-etal-2013-webqa}.

\subsubsection{Implementation Details}
We use the hyperparameters reported for the existing models whenever available. Implementation details are available in the Appendix~\ref{sec:hyper}.

\subsubsection{Baselines and Metrics}
To investigate the enhancement effects of thinking and planning on complex RAG tasks, we compare it with vanilla RAG, which performs only a single retrieval and generation process. 
We evaluate the naive generators of \textit{Qwen2.5}~\citep{qwen2.5} series instruction models and \textit{Llama3-8B-Instruct}~\citep{grattafiori2024llama3herdmodels}. In the retrieval phase, we employ \textit{multilingual-e5-base}~\citep{wang2024multilinguale5} as the retriever and utilize the widely used Wikipedia dump from December 2018 as the retrieval corpus~\citep{karpukhin-etal-2020-dense}.
The prompt of vanilla RAG are shown in the Appendix~\ref{sec:prompt_examples}. 
For iterative retrieval, we compare AirRAG with Iter-RetGen~\citep{shao-etal-2023-iterretgen}, Self-RAG~\citep{asai-2024-selfrag}, Auto-RAG~\citep{yu2024autorag}, and IterDRAG~\citep{yue2024iterdrag}. For agentic retrieval, we compare AirRAG with Search-o1~\citep{li2025searcho1}, ReSearch~\citep{chen2025research}, and DeepResearcher~\citep{zheng2025deepresearcher}.
To further explore RAG performance and inference computation scaling, we focus on a comparison with IterDRAG for a given budget on inference computation. For evaluation metrics, we report Exact Match (EM), F1 score (F1) and Accuracy (Acc) between the generated summary and gold answer, where accuracy measures whether the gold answer is covered in the generated answer.

\subsection{Main Results (RQ1)}\label{main_results}
We first evaluate the performance of AirRAG on various complex QA datasets. Table~\ref{tab:main-result} compares its accuracy and F1 scores with strong baselines based on LLMs of different scales. The optimal performance exhibits consistent gains as the LLMs scale up. For the \textit{Qwen2.5-7b-instruct} model, our approach achieves the best performance, even surpassing the trainable approaches. To further validate its effectiveness on large-scale reasoning models, we also conduct experiments on \textit{Qwen3-235B} in both the thinking mode and non-thinking mode. In thinking mode, our approach achieves state-of-the-art performance among all datasets.
In addition, to verify the robustness and generalization of AirRAG, Table~\ref{tab:main-result-llama} shows the performance on more diverse LLMs and datasets. 
We observe consistent improvements over vanilla RAG and existing iterative methods (more than 10\% on average). The significant boost over IterDRAG and Auto-RAG suggests that AirRAG explores more effective reasoning paths through the human-like thinking paradigm and tree-based search.

\begin{table*}[!t] 
 \small
 \centering
\resizebox{1.00\textwidth}{!}{
\begin{tabular}{lcccccccccccc} 
\toprule
 \multirow{2}{*}{\textbf{Method}}  & \multicolumn{2}{c}{\textbf{NQ}} & \multicolumn{2}{c}{\textbf{TriviaQA}} & \multicolumn{2}{c}{\textbf{HotpotQA}}  & \multicolumn{2}{c}{\textbf{MuSiQue}} & \multicolumn{2}{c}{\textbf{2Wiki}} & \multicolumn{2}{c}{\textbf{Average}} \\  
\cmidrule(l){2-13}  & F1 & Acc & F1 & Acc & F1 & Acc & F1 & Acc & F1 & Acc & F1& Acc\\ 
\hline
    \rowcolor[rgb]{0.9,0.9,0.9}
    \multicolumn{1}{l}
    {\textbf{\textit{Qwen2.5-7B}}} &  &  &  & &&&& &  &  &  &  \\
    \multicolumn{1}{l}{\text{ZeroShot QA}} & 38.4 & 37.4 & 57.7 & 56.3 & 36.1 & 34.8 & 9.1 & 7.5 & 45.0 & 44.1 & 37.3 & 36.0  \\
    \multicolumn{1}{l}{\text{Vanilla RAG}} & 57.8 & 53.9 & 70.1 & 66.3 & 61.3 & 56.9 & 13.5 & 8.3 & 45.9 & 42.8 & 49.7 & 45.6 \\
    \multicolumn{1}{l}{\text{IterDRAG}$^{*}$} & 58.3 & 54.1 & 73.5 & 69.1 & 65.3 & 60.7 & 18.3 & 13.0 & 51.8 & 47.0 & 53.4 & 48.8 \\
    \multicolumn{1}{l}{\text{Search-o1}$^{*}$} &57.8 & 54.3 & 72.6 & 69.8 &   57.3 & 54.2 & 20.2 & 18.5 &  56.9 & 53.6 &  53.0& 50.1 \\
    \multicolumn{1}{l}{\text{ReSearch}$^{*}$} & 61.3 &  59.6& \textbf{76.2} & \textbf{73.4} & 70.3 & 68.1 & 30.9 & 27.4 & 62.5& 61.0 & 60.2 & \textbf{57.9} \\
    \multicolumn{1}{l}{\text{DeepResearcher}} & \textbf{62.4} & \textbf{59.8} & 75.9 &  73.2 & 67.6 & 63.9 & \textbf{33.9 } & \textbf{27.8} & 65.4 & 62.3 &61.0 &57.4 \\
    \midrule
    \multicolumn{1}{l}{\text{AirRAG-Lite}} & 60.8 & 57.3 & 74.1 & 70.0 & 68.2 & 63.2 & 23.4 & 17.3 & 51.1 & 48.0 & 55.5 & 51.2 \\
    \multicolumn{1}{l}{\text{AirRAG}} & 60.4 & 56.2 & 74.3 & 70.1 & \textbf{74.9} & \textbf{70.0} & 30.3 & 23.6 & \textbf{65.7} & \textbf{63.2} & \textbf{61.1} & 56.6 \\
\hline
    \rowcolor[rgb]{0.9,0.9,0.9}
    \multicolumn{1}{l}
    {\textbf{\textit{Qwen2.5-14B}}} &  &  &  & &&&& &  &  &  &  \\
    \multicolumn{1}{l}{\text{ZeroShot QA}} & 47.1 & 46.3 & 70.7 & 69.5 & 42.5 & 41.3 & 13.5 & 12.1 & 48.2 & 47.3 & 44.4 & 43.3 \\
    \multicolumn{1}{l}{\text{Vanilla RAG}} & 62.0 & 58.2 & 75.2 & 71.2 & 67.0 & 62.0 & 20.0 & 14.4 & 52.0 & 49.4 & 55.2 & 51.0 \\
    \multicolumn{1}{l}{\text{IterDRAG}$^{*}$} & 58.5 & 53.9 & 76.4 & 72.3 & 71.8 & 66.2 & 23.9 & 17.2 & 57.4 & 54.1 & 57.6 & 52.7 \\
    \multicolumn{1}{l}{\text{Search-o1}$^{*}$} & 61.2 & 59.7&  74.3 & 72.6 & 71.4 & 67.4 & 23.2 & 20.4 &  58.1&55.8  & 57.6 & 55.2 \\
    \midrule
    \multicolumn{1}{l}{\text{AirRAG-Lite}} & 64.8 & 60.9 & \textbf{78.9} & \textbf{75.0} & 77.2 & 72.5 & 30.3 & 24.3 & 70.3 & 68.0 & 64.3 & 60.1 \\
    \multicolumn{1}{l}{\text{AirRAG}} & \textbf{66.2} & \textbf{62.1} & 78.1 & 73.8 & \textbf{79.9} & \textbf{75.3} & \textbf{36.0} & \textbf{31.9} & \textbf{70.4} & \textbf{68.7} & \textbf{66.1} & \textbf{62.4} \\
\hline
    \rowcolor[rgb]{0.9,0.9,0.9}
    \multicolumn{1}{l}
    {\textbf{\textit{Qwen2.5-32B}}} &  &  &  & &&&& &  &  &  &  \\
    \multicolumn{1}{l}{\text{ZeroShot QA}} & 46.8 & 45.9 & 69.8 & 68.8 & 42.9 & 41.8 & 11.1 & 9.6 & 48.1 & 47.2 & 43.7 & 42.7 \\
    \multicolumn{1}{l}{\text{Vanilla RAG}} & 60.6 & 56.7 & 75.4 & 71.6 & 66.5 & 62.5 & 19.5 & 13.8 & 51.8 & 49.6 & 54.7 & 50.8 \\
    \multicolumn{1}{l}{\text{IterDRAG}$^{*}$} & 61.1 & 56.7 & 76.9 & 73.0 & 72.0 & 67.3 & 23.3 & 17.9 & 58.2 & 55.4 & 58.3 & 54.1 \\
    \multicolumn{1}{l}{\text{Search-o1}$^{*}$} & 63.5&61.6 & 75.6& 72.8 & 72.8 & 68.4 & 30.2 & 27.3 & 60.5 & 58.9 & 60.5 &57.8  \\
    \multicolumn{1}{l}{\text{ReSearch}$^{*}$} & 63.8 &  61.2 & 74.6 &72.3 & 76.2 & 72.6 & \textbf{38.3} & \textbf{33.4} & 66.8 & 62.8 & 63.9 & 60.5 \\
    \midrule
    \multicolumn{1}{l}{\text{AirRAG-Lite}} & 65.6 & 61.9 & 75.8 & 71.6 & 79.3 & 74.8 & 33.3 & 32.5 & \textbf{72.0} & \textbf{70.8} & 65.2 & 62.3 \\
    \multicolumn{1}{l}{\text{AirRAG}} & \textbf{66.5} & \textbf{62.7 }& \textbf{78.9} & \textbf{74.6} & \textbf{81.1 }& \textbf{76.1} & 36.5 & 32.7 & 71.9 & 70.6 & \textbf{67.0} & \textbf{63.3} \\
\hline
    \rowcolor[rgb]{0.9,0.9,0.9}
    \multicolumn{1}{l}{\textbf{\textit{Qwen3-235B (non-thinking)}}} &  &  &  & &&&& &  &  &  &  \\
    \multicolumn{1}{l}{\text{ZeroShot QA}} & 64.1 & 63.7 & 77.1 & 76.3 & 53.9 & 52.9 & 17.1 & 15.7 & 56.8 & 56.1 & 53.8 & 52.9 \\
    \multicolumn{1}{l}{\text{Vanilla RAG}} & 66.0 & 65.3 & 78.0 & 76.2 & 69.2 & 67.7 & 20.8 & 19.1 & 53.3 & 52.2 & 57.4 & 56.1 \\
    \multicolumn{1}{l}{\text{IterDRAG}$^{*}$} & 65.7 & 63.3 & 78.8 & 76.9 & 75.5 & 69.7 & 32.2 & 25.2 & 64.8 & 60.8 & 63.4 & 59.2 \\
    \multicolumn{1}{l}{\text{Search-o1}$^{*}$} & 67.3 & 66.2 &  77.3& 76.4 & 75.9 & 73.8 & 36.7 & 34.2 & 71.3 & 68.2 & 65.7 & 63.7  \\
    \midrule
    \multicolumn{1}{l}{\text{AirRAG-Lite}} & \textbf{67.7} & \textbf{66.9} & 77.6 & 75.8 & 78.3 & 76.8 & 43.7 & 36.5 & 74.9 & 74.1 & 68.4 & 65.7 \\
    \multicolumn{1}{l}{\text{AirRAG}} & 66.4 & 65.6 & \textbf{79.1} & \textbf{77.3} & \textbf{79.6} & \textbf{78.1} & \textbf{47.2} & \textbf{40.0} & \textbf{76.2}& \textbf{75.5} & \textbf{69.7} & \textbf{67.3} \\
\hline
    \rowcolor[rgb]{0.9,0.9,0.9}
    \multicolumn{1}{l}{\textbf{\textit{Qwen3-235B (thinking)}}} &  &  &  & &&&& &  &  &  &  \\
    \multicolumn{1}{l}{\text{ZeroShot QA}} & 66.1 & 65.6 & 79.3 & 78.6 & 54.2 & 53.5 & 21.3 & 19.9 & 56.6 & 56.3 & 55.5 & 54.8 \\
    \multicolumn{1}{l}{\text{Vanilla RAG}} & 67.6 & 67.1 & 77.9 & 76.9 & 72.5 & 71.9 & 18.2 & 16.4 & 57.5 & 56.6 & 58.7 & 57.8 \\
    \multicolumn{1}{l}{\text{IterDRAG}$^{*}$} & 68.8 & 67.1 & 80.9 & 78.7 & 76.3 & 71.1 & 30.0 & 23.1 & 67.1 & 64.5 & 64.6 & 60.9 \\
    \multicolumn{1}{l}{\text{Search-o1}$^{*}$} & 67.2 & 66.4 &  78.1 & 77.5 & 75.2 & 72.4 & 33.4 & 28.9 & 69.2 & 66.3 & 64.6 & 62.3  \\
    \midrule
    \multicolumn{1}{l}{\text{AirRAG-Lite}} & 73.2 & 72.7 & 81.1 & 80.1 & \textbf{86.2} & \textbf{85.6} & 44.2 & 37.8 & 76.4 & 75.6 & 72.2 & 70.3 \\
    \multicolumn{1}{l}{\text{AirRAG}} & \textbf{74.3} & \textbf{72.8} & \textbf{81.4} & \textbf{80.1} & 84.7 & 84.0 & \textbf{47.5} & \textbf{40.3} & \textbf{76.8} & \textbf{76.2} & \textbf{72.9} & \textbf{70.7} \\
\bottomrule
\end{tabular}
}
 \caption{Overall evaluation results on the test sets of five datasets. * indicates the results reproduced by us. The best results for each model are in \textbf{bold}. The number of both rollouts and output sequences is set to 1. The number of documents for a single retrieval is set to 5.}
 \label{tab:main-result}
\end{table*}

\subsection{Inference Scaling for RAG (RQ2)}\label{sec:inference_scaling_exp}
\begin{figure*}[t]
\centering
\includegraphics[width=0.99\textwidth]{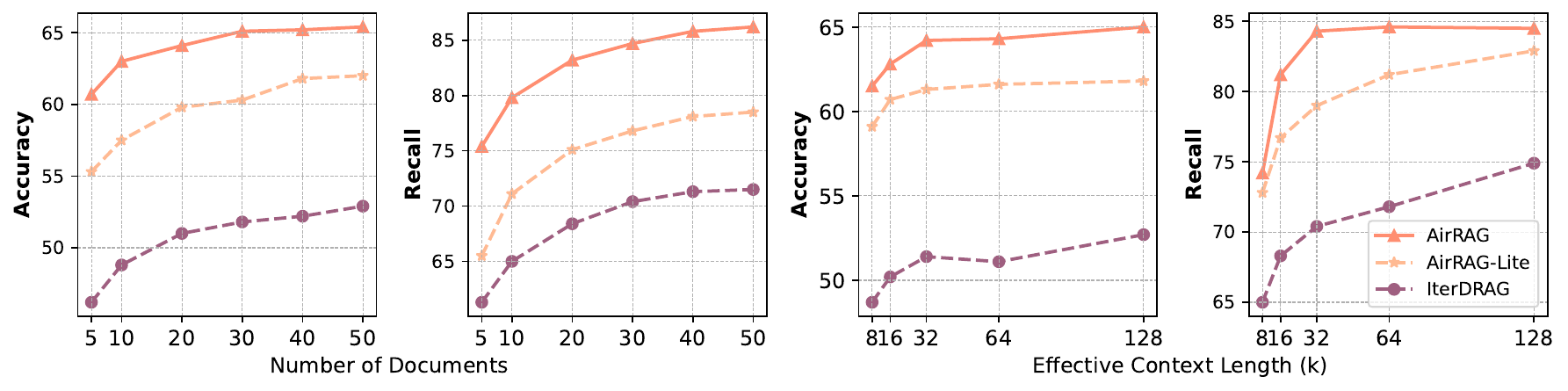}
\caption{Impact of the retrieved document number scaling (\textbf{Left}) and the maximum context length scaling (\textbf{Right}) on model performance (averaged Accuracy and Recall of three datasets). All methods show consistent performance improvements as the effective inference computation scales.}
\label{fig:nScaling}
\end{figure*}

\begin{figure}[t]
\centering
\includegraphics[width=0.48\textwidth]{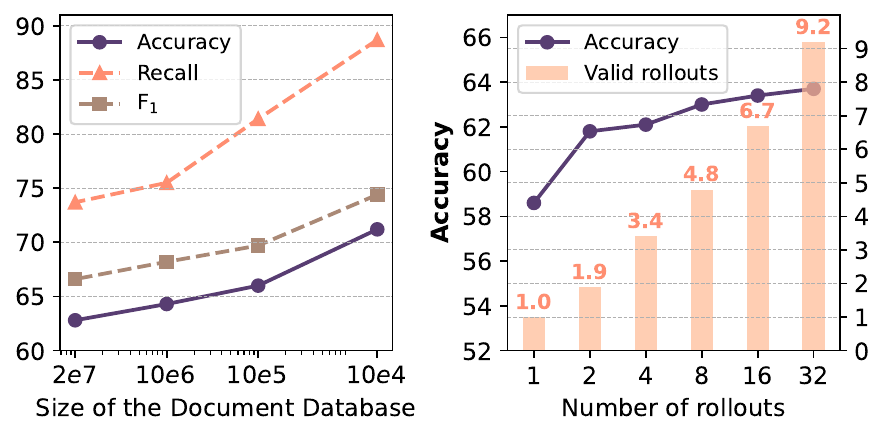}
\caption{\textbf{Left}: Performance comparison under different size of document database. A streamlined database can maintain a better performance. \textbf{Right}: Performance comparison in increasing the number of valid rollouts. Sampling a higher number of diverse reasoning paths consistently improves accuracy.}
\label{fig:DocSize_Rollout_Scaling}
\end{figure}

Inference computation scaling can enable LLMs to improve their output performance~\citep{snell2024scaling}. Self-consistency can also improve the robustness of the reasoning process~\citep{wang2023selfconsistency}. Therefore, we carry out a comprehensive experimental analysis on the inference computation scaling. Based on tree-based search and RAG scenario, there are multiple ways to optimize the use of inference computation resources. Specifically, we can adjust both the number of retrieved documents in a single retrieval and the effective context length in all iterations. The average performance of three datasets exhibits consistent gains in Figure~\ref{fig:nScaling}. 
In subsequent experiments, unless otherwise specified, the data presented represent the average performance across the HotpotQA, MuSiQue, and 2Wiki datasets.
In particular, the initial computation scaling brings significant performance improvements. In addition, the number of output sequences and rollouts in MCTS can expand the solution space and explore potential reasoning paths.
As shown in Figure~\ref{fig:intro}, the average performance increases with the number of output sequences per action, demonstrating the effectiveness of self-consistency. We also investigate the number of effective reasoning paths under different rollouts in Figure~\ref{fig:DocSize_Rollout_Scaling}. The performance improvement caused by the increase of effective reasoning paths in the early stage is relatively high. We provide additional dataset-specific results in Appendix~\ref{sec:additional_exp}.

\begin{table}[htbp]
 \small
 \centering
\begin{tabular} {@{}ccc @{}} 
\toprule
\multirow{2}{*}{Method}  & \multicolumn{2}{c}{Average} \\  
\cmidrule(l){2-3}  & F1 & Acc\\ 
\midrule

\multicolumn{1}{l}{\text{Vanilla RAG}} & 47.0 & 43.2 \\
\multicolumn{1}{l}{\text{IterDRAG}} & 49.8 & 45.9  \\

\multicolumn{1}{l}{\text{AirRAG}} &  &  \\
\multicolumn{1}{l}{{ + $n_{\text{all}}$=1}} & 62.9 & 58.8 \\
\multicolumn{1}{l}{{ + $n_{\text{all}}$=3}} & \underline{63.4} & \underline{62.1} \\
\multicolumn{1}{l}{{ + $n_{a_1,a_4}$=3, n$_{a_2,a_3,a_5}$=1}} & 63.2 & 62.0 \\
\multicolumn{1}{l}{{ + $n_{a_1,a_4}$=3, $n_{a_2,a_3,a_5}$=1, $q_{div}$=1.0}} & \textbf{65.1} & \textbf{63.9} \\
\bottomrule
\end{tabular}
 \caption{Performance comparison with different computationally optimal strategies on the HotpotQA, MuSiQue and 2Wiki datasets. $n_{a_i}$ denotes the number of output sequences of the action $a_i$ in a single extension. $q_{div}$ indicates that setting top-$p$ to 1.0 and temperature to 1.0 for query-related actions, i.e. SAY and QT, increases the diversity of reasoning. The default sampling parameters top-$p$, top-$k$ and temperature are set to 0.8, 50 and 0.7 respectively. Rational sampling strategies further improve performance across multiple datasets.}
 \label{tab:main-result-n}
\end{table}

\subsection{Ablation Studies}\label{ablation}
\noindent\textbf{Effect of Computationally Optimal Strategies (RQ3)}.
Extensive experiments show that the outputs of certain actions (e.g., RA, DA and SA) are almost consistent when performing multiple generations. Therefore, we only increase the number of output sequences (short for $n$) for the remaining actions (e.g., SAY and QT), which reduces invalid inference computation while maintaining good results. This also reflects that this kind of reasoning action, which effectively activates the creativity of LLMs,
requires more diversified sampling strategies. We adjust the sampling parameters (top-$p$=1.0 and temperature=1.0) to improve the diversity of the model output. The complete experimental results in Table~\ref{tab:main-result-n} show that the diversity of key actions can significantly improve performance.

From the aforementioned experiments, it is observed that the recall and accuracy of model are linearly correlated. Intuitively, the size of document database is also related to the recall score. By reducing the scale of the document database, we find a gradual improvement in model performance (shown in Figure~\ref{fig:DocSize_Rollout_Scaling}). This observation provides experimental evidence for effective database partitioning in practical application.

\begin{table*}[htbp]
 \centering
 \resizebox{0.75\textwidth}{!}{
\begin{tabular} {@{}clcccccccc @{}}
\toprule
\text{$L_{\max}$} & \text{Method}  & \text{database\_size} & \text{retrieval\_time} & \text{retrieval\_number} & \text{e2e} \\  
\midrule
\multirow{4}{*}{8k} &
\multicolumn{1}{l}{\text{Vanilla RAG}} & 100w & 0.900 & 1.00 & 3.828 \\
 &
\multicolumn{1}{l}{\text{IterDRAG}} & 100w & 1.833 & 2.04 & 6.703 \\
 &
\multicolumn{1}{l}{\text{AirRAG-Lite}} & 100w & 2.205 & 2.25 & 8.482 \\
 &
\multicolumn{1}{l}{\text{AirRAG}} & 100w & 3.453 & 3.84 &  12.752  \\
\bottomrule
\end{tabular}
}
 \caption{
 Performance analysis of inference efficiency. $L_{\max}$ denotes the maximum number of input tokens across all rollouts. The retrieved database contains approximately one million documents. e2e and retrieval\_time denote the average total time for a single question-answering process and the time spent on retrieval respectively, measured in seconds. Other inference configurations remain consistent with those in Table~\ref{tab:scaling-main-result}.}
 \label{tab:inference_efficiency}
\end{table*}

\noindent\textbf{Effect of Verification Methods (RQ4)}.
The larger search space also generates more candidate reasoning trajectories. Therefore, how to select the optimal trajectory is crucial for the final performance. We compare multiple verification methods with the average scores of all candidates in Figure~\ref{fig:verify}. These two self-consistency verification methods are always slightly better than the average score, but they are not nearly as good as the SA and QwenRM methods. The SA method uses the LLM to further refine the final answer from all candidate rollouts, which is simple and effective. Finally, the reward model achieves the most competitive results due to the introduction of supervised information on key intermediate reasoning steps in the training process. However, collecting process-supervised training samples requires high computational costs and high-quality raw data. In the practical application scenario, we can choose the appropriate method while balancing efficiency and effectiveness.

\subsection{Inference Efficiency and Qualitative Analysis (RQ5)}\label{case_study}
Given the inherently large search space for the tree-based search, we design computational optimization strategies for different actions to avoid redundant and inefficient expansions, as shown in Table~\ref{tab:main-result-n}. Furthermore, in Section~\ref{sec:3.2.2}, we propose pruning strategies for state nodes and reasoning paths. These optimizations significantly reduce inefficient LLM inference and repetitive path exploration. In practical applications, we can select appropriate configuration parameters such as rollout, $n$, and $L_{\max}$ based on computational resource budgets and time constraints, ensuring effectiveness while achieving inference efficiency comparable to current mainstream iterative RAG approaches. We analyze the average inference efficiency per sample on the HotpotQA dataset in Table~\ref{tab:inference_efficiency}. By combining these results with those reported in Table~\ref{tab:scaling-main-result} and Table~\ref{tab:main-result}, we are able to analyze the trade-offs between computational cost and performance. 

To make it easier to understand why our proposed AirRAG works, we present a qualitative analysis in MuSiQue. 
Existing iterative methods are often trapped in a single solution space when confronted with complex tasks. As illustrated in Figure~\ref{fig:iterative_ambiguous_query}, these iterative methods exhibit a key limitation that insufficient or ambiguous retrieval context can lead to repetitive follow-up queries until it reaches the predefined maximum depth of iterations. This inefficient iteration results in high computational cost and incorrect answer.
In contrast, AirRAG designs efficient reasoning actions to achieve autonomous planning and reasoning. As shown in Figure~\ref{fig:air_rag_case}, the SAY action decomposes the original query into a more rational sequence of sub-queries, and then the combination of RA and QT ensures the accuracy of the intermediate reasoning step. We eventually leverage the efficient reasoning trajectory to obtain the correct answer.

\section{Conclusions}
In this paper, we propose AirRAG, a novel RAG approach to fully leverage the planning and reasoning capabilities of LLMs. AirRAG designs an efficient action space for the controllable reasoning generation. We also introduce Monte Carlo Tree Search to expand the solution space. Meanwhile, by employing the tree-based search and self-consistency verification, we explore potential reasoning paths and achieve comprehensive inference computation scaling. In addition, computationally optimal strategies are used to apply more computation to key actions, leading to further performance improvements. Experimental results on diverse QA datasets demonstrate the significant superiority of AirRAG over other methods designed for complex deep search scenarios.

\section*{Limitations} 
Although our model achieves competitive performance in various RAG tasks, there are some limitations that can be improved. The current optimal computation allocation strategy is derived from sufficient experiments. We can consider designing an automated policy model to implement the trade-off between computational cost and performance. Despite great efforts in the inference scaling of RAG, the experimental analysis may be limited due to the massive computational cost of tree-based search approaches. We will explore more complex reasoning tasks to verify the robustness and effectiveness of our approach. In addition, the large search space also brings more noise information, so we will further investigate the reward model or strategy to explore a better reasoning path.


\bibliography{custom}

\begin{thebibliography}{42}
\providecommand{\natexlab}[1]{#1}

\bibitem[{Asai et~al.(2024)Asai, Wu, Wang, Sil, and Hajishirzi}]{asai-2024-selfrag}
Akari Asai, Zeqiu Wu, Yizhong Wang, Avirup Sil, and Hannaneh Hajishirzi. 2024.
\newblock \href {https://openreview.net/forum?id=hSyW5go0v8} {Self-{RAG}: Learning to retrieve, generate, and critique through self-reflection}.
\newblock In \emph{The Twelfth International Conference on Learning Representations}.

\bibitem[{Berant et~al.(2013)Berant, Chou, Frostig, and Liang}]{berant-etal-2013-webqa}
Jonathan Berant, Andrew Chou, Roy Frostig, and Percy Liang. 2013.
\newblock \href {https://aclanthology.org/D13-1160/} {Semantic parsing on {F}reebase from question-answer pairs}.
\newblock In \emph{Proceedings of the 2013 Conference on Empirical Methods in Natural Language Processing}, pages 1533--1544, Seattle, Washington, USA. Association for Computational Linguistics.

\bibitem[{Besta et~al.(2024)Besta, Blach, Kubicek, Gerstenberger, Gianinazzi, Gajda, Lehmann, Podstawski, Niewiadomski, Nyczyk, and Hoefler}]{besta2024got}
Maciej Besta, Nils Blach, Ales Kubicek, Robert Gerstenberger, Lukas Gianinazzi, Joanna Gajda, Tomasz Lehmann, Micha{\l} Podstawski, Hubert Niewiadomski, Piotr Nyczyk, and Torsten Hoefler. 2024.
\newblock \href {https://doi.org/10.1609/aaai.v38i16.29720} {{Graph of Thoughts: Solving Elaborate Problems with Large Language Models}}.
\newblock \emph{Proceedings of the AAAI Conference on Artificial Intelligence}, 38(16):17682--17690.

\bibitem[{Chen et~al.(2024)Chen, Liao, Li, and Fan}]{chen2024alphamath}
Guoxin Chen, Minpeng Liao, Chengxi Li, and Kai Fan. 2024.
\newblock \href {https://openreview.net/forum?id=VaXnxQ3UKo} {Alphamath almost zero: Process supervision without process}.
\newblock In \emph{The Thirty-eighth Annual Conference on Neural Information Processing Systems}.

\bibitem[{Chen et~al.(2025)Chen, Li, Sun, Zhou, Zhu, Wang, Pan, Zhang, Chen, Yang, Zhou, and Chen}]{chen2025research}
Mingyang Chen, Tianpeng Li, Haoze Sun, Yijie Zhou, Chenzheng Zhu, Haofen Wang, Jeff~Z. Pan, Wen Zhang, Huajun Chen, Fan Yang, Zenan Zhou, and Weipeng Chen. 2025.
\newblock \href {https://arxiv.org/abs/2503.19470} {Research: Learning to reason with search for llms via reinforcement learning}.
\newblock \emph{Preprint}, arXiv:2503.19470.

\bibitem[{Gao et~al.(2023)Gao, Ma, Lin, and Callan}]{gao-etal-2023-precise-hyde}
Luyu Gao, Xueguang Ma, Jimmy Lin, and Jamie Callan. 2023.
\newblock \href {https://doi.org/10.18653/v1/2023.acl-long.99} {Precise zero-shot dense retrieval without relevance labels}.
\newblock In \emph{Proceedings of the 61st Annual Meeting of the Association for Computational Linguistics (Volume 1: Long Papers)}, pages 1762--1777, Toronto, Canada. Association for Computational Linguistics.

\bibitem[{Grattafiori et~al.(2024)Grattafiori, Dubey, Jauhri, Pandey, Kadian, Al-Dahle, Letman, Mathur, Schelten, Vaughan, Yang, and et~al.}]{grattafiori2024llama3herdmodels}
Aaron Grattafiori, Abhimanyu Dubey, Abhinav Jauhri, Abhinav Pandey, Abhishek Kadian, Ahmad Al-Dahle, Aiesha Letman, Akhil Mathur, Alan Schelten, Alex Vaughan, Amy Yang, and et~al. 2024.
\newblock \href {https://arxiv.org/abs/2407.21783} {The llama 3 herd of models}.
\newblock \emph{Preprint}, arXiv:2407.21783.

\bibitem[{Hendrycks et~al.(2021)Hendrycks, Burns, Kadavath, Arora, Basart, Tang, Song, and Steinhardt}]{hendrycks2021MATH}
Dan Hendrycks, Collin Burns, Saurav Kadavath, Akul Arora, Steven Basart, Eric Tang, Dawn Song, and Jacob Steinhardt. 2021.
\newblock \href {https://openreview.net/forum?id=7Bywt2mQsCe} {Measuring mathematical problem solving with the {MATH} dataset}.
\newblock In \emph{Thirty-fifth Conference on Neural Information Processing Systems Datasets and Benchmarks Track (Round 2)}.

\bibitem[{Ho et~al.(2020)Ho, Duong~Nguyen, Sugawara, and Aizawa}]{ho-etal-2020-2wiki}
Xanh Ho, Anh-Khoa Duong~Nguyen, Saku Sugawara, and Akiko Aizawa. 2020.
\newblock \href {https://doi.org/10.18653/v1/2020.coling-main.580} {Constructing a multi-hop {QA} dataset for comprehensive evaluation of reasoning steps}.
\newblock In \emph{Proceedings of the 28th International Conference on Computational Linguistics}, pages 6609--6625, Barcelona, Spain (Online). International Committee on Computational Linguistics.

\bibitem[{Jaffe et~al.(2023)Jaffe, Poldrack, Schafer, and Bissett}]{jaffe2023modelling}
Paul~I Jaffe, Russell~A Poldrack, Robert~J Schafer, and Patrick~G Bissett. 2023.
\newblock Modelling human behaviour in cognitive tasks with latent dynamical systems.
\newblock \emph{Nature Human Behaviour}, 7(6):986--1000.

\bibitem[{Jeong et~al.(2024)Jeong, Baek, Cho, Hwang, and Park}]{jeong-etal-2024-adaptive}
Soyeong Jeong, Jinheon Baek, Sukmin Cho, Sung~Ju Hwang, and Jong Park. 2024.
\newblock \href {https://doi.org/10.18653/v1/2024.naacl-long.389} {Adaptive-{RAG}: Learning to adapt retrieval-augmented large language models through question complexity}.
\newblock In \emph{Proceedings of the 2024 Conference of the North American Chapter of the Association for Computational Linguistics: Human Language Technologies (Volume 1: Long Papers)}, pages 7036--7050, Mexico City, Mexico. Association for Computational Linguistics.

\bibitem[{Jiang et~al.(2023)Jiang, Xu, Gao, Sun, Liu, Dwivedi-Yu, Yang, Callan, and Neubig}]{jiang-etal-2023-active}
Zhengbao Jiang, Frank Xu, Luyu Gao, Zhiqing Sun, Qian Liu, Jane Dwivedi-Yu, Yiming Yang, Jamie Callan, and Graham Neubig. 2023.
\newblock \href {https://doi.org/10.18653/v1/2023.emnlp-main.495} {Active retrieval augmented generation}.
\newblock In \emph{Proceedings of the 2023 Conference on Empirical Methods in Natural Language Processing}, pages 7969--7992, Singapore. Association for Computational Linguistics.

\bibitem[{Jin et~al.(2024)Jin, Zhu, Yang, Zhang, and Dou}]{FlashRAG}
Jiajie Jin, Yutao Zhu, Xinyu Yang, Chenghao Zhang, and Zhicheng Dou. 2024.
\newblock \href {https://arxiv.org/abs/2405.13576} {Flashrag: A modular toolkit for efficient retrieval-augmented generation research}.
\newblock \emph{CoRR}, abs/2405.13576.

\bibitem[{Joshi et~al.(2017)Joshi, Choi, Weld, and Zettlemoyer}]{joshi-etal-2017-triviaqa}
Mandar Joshi, Eunsol Choi, Daniel Weld, and Luke Zettlemoyer. 2017.
\newblock \href {https://doi.org/10.18653/v1/P17-1147} {{T}rivia{QA}: A large scale distantly supervised challenge dataset for reading comprehension}.
\newblock In \emph{Proceedings of the 55th Annual Meeting of the Association for Computational Linguistics (Volume 1: Long Papers)}, pages 1601--1611, Vancouver, Canada. Association for Computational Linguistics.

\bibitem[{Kandpal et~al.(2023)Kandpal, Deng, Roberts, Wallace, and Raffel}]{pmlr-v202-kandpal23a}
Nikhil Kandpal, Haikang Deng, Adam Roberts, Eric Wallace, and Colin Raffel. 2023.
\newblock \href {https://proceedings.mlr.press/v202/kandpal23a.html} {Large language models struggle to learn long-tail knowledge}.
\newblock In \emph{Proceedings of the 40th International Conference on Machine Learning}, volume 202 of \emph{Proceedings of Machine Learning Research}, pages 15696--15707. PMLR.

\bibitem[{Karpukhin et~al.(2020)Karpukhin, Oguz, Min, Lewis, Wu, Edunov, Chen, and Yih}]{karpukhin-etal-2020-dense}
Vladimir Karpukhin, Barlas Oguz, Sewon Min, Patrick Lewis, Ledell Wu, Sergey Edunov, Danqi Chen, and Wen-tau Yih. 2020.
\newblock \href {https://doi.org/10.18653/v1/2020.emnlp-main.550} {Dense passage retrieval for open-domain question answering}.
\newblock In \emph{Proceedings of the 2020 Conference on Empirical Methods in Natural Language Processing (EMNLP)}, pages 6769--6781, Online. Association for Computational Linguistics.

\bibitem[{Kocsis and Szepesv{\'a}ri(2006)}]{UCT2006}
Levente Kocsis and Csaba Szepesv{\'a}ri. 2006.
\newblock Bandit based monte-carlo planning.
\newblock In \emph{Machine Learning: ECML 2006}, pages 282--293, Berlin, Heidelberg. Springer Berlin Heidelberg.

\bibitem[{Kwiatkowski et~al.(2019)Kwiatkowski, Palomaki, Redfield, Collins, Parikh, Alberti, Epstein, Polosukhin, Devlin, Lee, Toutanova, Jones, Kelcey, Chang, Dai, Uszkoreit, Le, and Petrov}]{kwiatkowski-etal-2019-naturalQA}
Tom Kwiatkowski, Jennimaria Palomaki, Olivia Redfield, Michael Collins, Ankur Parikh, Chris Alberti, Danielle Epstein, Illia Polosukhin, Jacob Devlin, Kenton Lee, Kristina Toutanova, Llion Jones, Matthew Kelcey, Ming-Wei Chang, Andrew~M. Dai, Jakob Uszkoreit, Quoc Le, and Slav Petrov. 2019.
\newblock \href {https://doi.org/10.1162/tacl_a_00276} {Natural questions: A benchmark for question answering research}.
\newblock \emph{Transactions of the Association for Computational Linguistics}, 7:452--466.

\bibitem[{Li et~al.(2025)Li, Dong, Jin, Zhang, Zhou, Zhu, Zhang, and Dou}]{li2025searcho1}
Xiaoxi Li, Guanting Dong, Jiajie Jin, Yuyao Zhang, Yujia Zhou, Yutao Zhu, Peitian Zhang, and Zhicheng Dou. 2025.
\newblock \href {https://arxiv.org/abs/2501.05366} {Search-o1: Agentic search-enhanced large reasoning models}.
\newblock \emph{Preprint}, arXiv:2501.05366.

\bibitem[{Ma et~al.(2023)Ma, Gong, He, Zhao, and Duan}]{ma-etal-2023-query}
Xinbei Ma, Yeyun Gong, Pengcheng He, Hai Zhao, and Nan Duan. 2023.
\newblock \href {https://doi.org/10.18653/v1/2023.emnlp-main.322} {Query rewriting in retrieval-augmented large language models}.
\newblock In \emph{Proceedings of the 2023 Conference on Empirical Methods in Natural Language Processing}, pages 5303--5315, Singapore. Association for Computational Linguistics.

\bibitem[{Mallen et~al.(2023)Mallen, Asai, Zhong, Das, Khashabi, and Hajishirzi}]{mallen-etal-2023-popqa}
Alex Mallen, Akari Asai, Victor Zhong, Rajarshi Das, Daniel Khashabi, and Hannaneh Hajishirzi. 2023.
\newblock \href {https://doi.org/10.18653/v1/2023.acl-long.546} {When not to trust language models: Investigating effectiveness of parametric and non-parametric memories}.
\newblock In \emph{Proceedings of the 61st Annual Meeting of the Association for Computational Linguistics (Volume 1: Long Papers)}, pages 9802--9822, Toronto, Canada. Association for Computational Linguistics.

\bibitem[{Nakano et~al.(2022)Nakano, Hilton, Balaji, Wu, Ouyang, Kim, Hesse, Jain, Kosaraju, Saunders, Jiang, Cobbe, Eloundou, Krueger, Button, Knight, Chess, and Schulman}]{nakano-2022webgpt}
Reiichiro Nakano, Jacob Hilton, Suchir Balaji, Jeff Wu, Long Ouyang, Christina Kim, Christopher Hesse, Shantanu Jain, Vineet Kosaraju, William Saunders, Xu~Jiang, Karl Cobbe, Tyna Eloundou, Gretchen Krueger, Kevin Button, Matthew Knight, Benjamin Chess, and John Schulman. 2022.
\newblock \href {https://arxiv.org/abs/2112.09332} {Webgpt: Browser-assisted question-answering with human feedback}.
\newblock \emph{Preprint}, arXiv:2112.09332.

\bibitem[{Qi et~al.(2024)Qi, Ma, Xu, Zhang, Yang, and Yang}]{qi2024rstar}
Zhenting Qi, Mingyuan Ma, Jiahang Xu, Li~Lyna Zhang, Fan Yang, and Mao Yang. 2024.
\newblock \href {https://arxiv.org/abs/2408.06195} {Mutual reasoning makes smaller llms stronger problem-solvers}.
\newblock \emph{Preprint}, arXiv:2408.06195.

\bibitem[{Schick et~al.(2023)Schick, Dwivedi-Yu, Dessi, Raileanu, Lomeli, Hambro, Zettlemoyer, Cancedda, and Scialom}]{schick-2023-toolformer}
Timo Schick, Jane Dwivedi-Yu, Roberto Dessi, Roberta Raileanu, Maria Lomeli, Eric Hambro, Luke Zettlemoyer, Nicola Cancedda, and Thomas Scialom. 2023.
\newblock \href {https://openreview.net/forum?id=Yacmpz84TH} {Toolformer: Language models can teach themselves to use tools}.
\newblock In \emph{Thirty-seventh Conference on Neural Information Processing Systems}.

\bibitem[{Shao et~al.(2023)Shao, Gong, Shen, Huang, Duan, and Chen}]{shao-etal-2023-iterretgen}
Zhihong Shao, Yeyun Gong, Yelong Shen, Minlie Huang, Nan Duan, and Weizhu Chen. 2023.
\newblock \href {https://doi.org/10.18653/v1/2023.findings-emnlp.620} {Enhancing retrieval-augmented large language models with iterative retrieval-generation synergy}.
\newblock In \emph{Findings of the Association for Computational Linguistics: EMNLP 2023}, pages 9248--9274, Singapore. Association for Computational Linguistics.

\bibitem[{Shi et~al.(2023)Shi, Chen, Misra, Scales, Dohan, Chi, Sch\"{a}rli, and Zhou}]{pmlr-v202-shi23a}
Freda Shi, Xinyun Chen, Kanishka Misra, Nathan Scales, David Dohan, Ed~H. Chi, Nathanael Sch\"{a}rli, and Denny Zhou. 2023.
\newblock \href {https://proceedings.mlr.press/v202/shi23a.html} {Large language models can be easily distracted by irrelevant context}.
\newblock In \emph{Proceedings of the 40th International Conference on Machine Learning}, volume 202 of \emph{Proceedings of Machine Learning Research}, pages 31210--31227. PMLR.

\bibitem[{Silver et~al.(2017)Silver, Hubert, Schrittwieser, Antonoglou, Lai, Guez, Lanctot, Sifre, Kumaran, Graepel et~al.}]{silver2017mastering}
David Silver, Thomas Hubert, Julian Schrittwieser, Ioannis Antonoglou, Matthew Lai, Arthur Guez, Marc Lanctot, Laurent Sifre, Dharshan Kumaran, Thore Graepel, et~al. 2017.
\newblock Mastering chess and shogi by self-play with a general reinforcement learning algorithm.
\newblock \emph{arXiv preprint arXiv:1712.01815}.

\bibitem[{Snell et~al.(2024)Snell, Lee, Xu, and Kumar}]{snell2024scaling}
Charlie Snell, Jaehoon Lee, Kelvin Xu, and Aviral Kumar. 2024.
\newblock \href {https://arxiv.org/abs/2408.03314} {Scaling llm test-time compute optimally can be more effective than scaling model parameters}.
\newblock \emph{Preprint}, arXiv:2408.03314.

\bibitem[{Trivedi et~al.(2022)Trivedi, Balasubramanian, Khot, and Sabharwal}]{trivedi-etal-2022-MuSiQue}
Harsh Trivedi, Niranjan Balasubramanian, Tushar Khot, and Ashish Sabharwal. 2022.
\newblock \href {https://doi.org/10.1162/tacl_a_00475} {{M}u{S}i{Q}ue: Multihop questions via single-hop question composition}.
\newblock \emph{Transactions of the Association for Computational Linguistics}, 10:539--554.

\bibitem[{Trivedi et~al.(2023)Trivedi, Balasubramanian, Khot, and Sabharwal}]{trivedi-etal-2023-ircot}
Harsh Trivedi, Niranjan Balasubramanian, Tushar Khot, and Ashish Sabharwal. 2023.
\newblock \href {https://doi.org/10.18653/v1/2023.acl-long.557} {Interleaving retrieval with chain-of-thought reasoning for knowledge-intensive multi-step questions}.
\newblock In \emph{Proceedings of the 61st Annual Meeting of the Association for Computational Linguistics (Volume 1: Long Papers)}, pages 10014--10037, Toronto, Canada. Association for Computational Linguistics.

\bibitem[{Wang et~al.(2024)Wang, Yang, Huang, Yang, Majumder, and Wei}]{wang2024multilinguale5}
Liang Wang, Nan Yang, Xiaolong Huang, Linjun Yang, Rangan Majumder, and Furu Wei. 2024.
\newblock Multilingual e5 text embeddings: A technical report.
\newblock \emph{arXiv preprint arXiv:2402.05672}.

\bibitem[{Wang et~al.(2023)Wang, Wei, Schuurmans, Le, Chi, Narang, Chowdhery, and Zhou}]{wang2023selfconsistency}
Xuezhi Wang, Jason Wei, Dale Schuurmans, Quoc~V Le, Ed~H. Chi, Sharan Narang, Aakanksha Chowdhery, and Denny Zhou. 2023.
\newblock \href {https://openreview.net/forum?id=1PL1NIMMrw} {Self-consistency improves chain of thought reasoning in language models}.
\newblock In \emph{The Eleventh International Conference on Learning Representations}.

\bibitem[{Yan et~al.(2024)Yan, Gu, Zhu, and Ling}]{yan2024corrective}
Shi-Qi Yan, Jia-Chen Gu, Yun Zhu, and Zhen-Hua Ling. 2024.
\newblock Corrective retrieval augmented generation.
\newblock \emph{arXiv preprint arXiv:2401.15884}.

\bibitem[{Yang et~al.(2024)Yang, Yang, Zhang, Hui, Zheng, Yu, Li, Liu, Huang, Wei, Lin, Yang, Tu, Zhang, Yang, Yang, Zhou, Lin, Dang, Lu, Bao, Yang, Yu, Li, Xue, Zhang, Zhu, Men, Lin, Li, Xia, Ren, Ren, Fan, Su, Zhang, Wan, Liu, Cui, Zhang, and Qiu}]{qwen2.5}
An~Yang, Baosong Yang, Beichen Zhang, Binyuan Hui, Bo~Zheng, Bowen Yu, Chengyuan Li, Dayiheng Liu, Fei Huang, Haoran Wei, Huan Lin, Jian Yang, Jianhong Tu, Jianwei Zhang, Jianxin Yang, Jiaxi Yang, Jingren Zhou, Junyang Lin, Kai Dang, Keming Lu, Keqin Bao, Kexin Yang, Le~Yu, Mei Li, Mingfeng Xue, Pei Zhang, Qin Zhu, Rui Men, Runji Lin, Tianhao Li, Tingyu Xia, Xingzhang Ren, Xuancheng Ren, Yang Fan, Yang Su, Yichang Zhang, Yu~Wan, Yuqiong Liu, Zeyu Cui, Zhenru Zhang, and Zihan Qiu. 2024.
\newblock Qwen2.5 technical report.
\newblock \emph{arXiv preprint arXiv:2412.15115}.

\bibitem[{Yang et~al.(2018)Yang, Qi, Zhang, Bengio, Cohen, Salakhutdinov, and Manning}]{yang-etal-2018-hotpotqa}
Zhilin Yang, Peng Qi, Saizheng Zhang, Yoshua Bengio, William Cohen, Ruslan Salakhutdinov, and Christopher~D. Manning. 2018.
\newblock \href {https://doi.org/10.18653/v1/D18-1259} {{H}otpot{QA}: A dataset for diverse, explainable multi-hop question answering}.
\newblock In \emph{Proceedings of the 2018 Conference on Empirical Methods in Natural Language Processing}, pages 2369--2380, Brussels, Belgium. Association for Computational Linguistics.

\bibitem[{Yao et~al.(2023)Yao, Yu, Zhao, Shafran, Griffiths, Cao, and Narasimhan}]{yao2023ToT}
Shunyu Yao, Dian Yu, Jeffrey Zhao, Izhak Shafran, Thomas~L. Griffiths, Yuan Cao, and Karthik~R Narasimhan. 2023.
\newblock \href {https://openreview.net/forum?id=5Xc1ecxO1h} {Tree of thoughts: Deliberate problem solving with large language models}.
\newblock In \emph{Thirty-seventh Conference on Neural Information Processing Systems}.

\bibitem[{Yu et~al.(2024)Yu, Zhang, and Feng}]{yu2024autorag}
Tian Yu, Shaolei Zhang, and Yang Feng. 2024.
\newblock \href {https://arxiv.org/abs/2411.19443} {Auto-rag: Autonomous retrieval-augmented generation for large language models}.
\newblock \emph{Preprint}, arXiv:2411.19443.

\bibitem[{Yue et~al.(2024)Yue, Zhuang, Bai, Hui, Jagerman, Zeng, Qin, Wang, Wang, and Bendersky}]{yue2024iterdrag}
Zhenrui Yue, Honglei Zhuang, Aijun Bai, Kai Hui, Rolf Jagerman, Hansi Zeng, Zhen Qin, Dong Wang, Xuanhui Wang, and Michael Bendersky. 2024.
\newblock \href {https://arxiv.org/abs/2410.04343} {Inference scaling for long-context retrieval augmented generation}.
\newblock \emph{Preprint}, arXiv:2410.04343.

\bibitem[{Zhang et~al.(2024)Zhang, Wu, Lei, Che, Li, Xie, Huang, Zhang, Pavone, Li, Ouyang, and Zhou}]{zhang2024llamaberry}
Di~Zhang, Jianbo Wu, Jingdi Lei, Tong Che, Jiatong Li, Tong Xie, Xiaoshui Huang, Shufei Zhang, Marco Pavone, Yuqiang Li, Wanli Ouyang, and Dongzhan Zhou. 2024.
\newblock \href {https://arxiv.org/abs/2410.02884} {Llama-berry: Pairwise optimization for o1-like olympiad-level mathematical reasoning}.
\newblock \emph{Preprint}, arXiv:2410.02884.

\bibitem[{Zheng et~al.(2024)Zheng, Mishra, Chen, Cheng, Chi, Le, and Zhou}]{zheng2024takeastepback}
Huaixiu~Steven Zheng, Swaroop Mishra, Xinyun Chen, Heng-Tze Cheng, Ed~H. Chi, Quoc~V Le, and Denny Zhou. 2024.
\newblock \href {https://openreview.net/forum?id=3bq3jsvcQ1} {Take a step back: Evoking reasoning via abstraction in large language models}.
\newblock In \emph{The Twelfth International Conference on Learning Representations}.

\bibitem[{Zheng et~al.(2025)Zheng, Fu, Hu, Cai, Ye, Lu, and Liu}]{zheng2025deepresearcher}
Yuxiang Zheng, Dayuan Fu, Xiangkun Hu, Xiaojie Cai, Lyumanshan Ye, Pengrui Lu, and Pengfei Liu. 2025.
\newblock \href {https://arxiv.org/abs/2504.03160} {Deepresearcher: Scaling deep research via reinforcement learning in real-world environments}.
\newblock \emph{Preprint}, arXiv:2504.03160.

\bibitem[{Zhou et~al.(2023)Zhou, Sch{\"a}rli, Hou, Wei, Scales, Wang, Schuurmans, Cui, Bousquet, Le, and Chi}]{zhou2023leasttomost}
Denny Zhou, Nathanael Sch{\"a}rli, Le~Hou, Jason Wei, Nathan Scales, Xuezhi Wang, Dale Schuurmans, Claire Cui, Olivier Bousquet, Quoc~V Le, and Ed~H. Chi. 2023.
\newblock \href {https://openreview.net/forum?id=WZH7099tgfM} {Least-to-most prompting enables complex reasoning in large language models}.
\newblock In \emph{The Eleventh International Conference on Learning Representations}.

\end{thebibliography}

\appendix
\clearpage
\section{Implementation Details}\label{sec:hyper}
\begin{figure*}[t]
\centering
\includegraphics[width=0.99\textwidth]{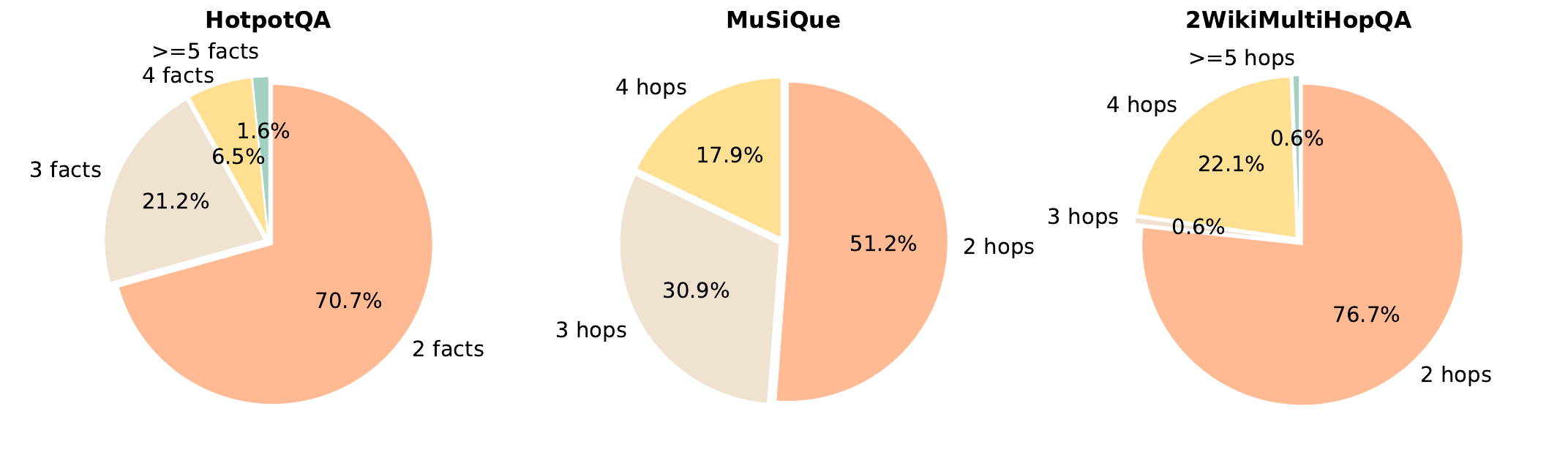}
\caption{Overview of the distribution of query complexity over three multi-hop QA datasets.}
\label{fig:DatasetHops}
\end{figure*}

For evaluation, we randomly select 1,000 samples from the whole validation sets of each dataset as our final test set, with a fixed random seed 0. To better understand the complexity of multi-hop reasoning in these datasets, we analyze the hop distribution of the HotpotQA, MuSiQue, and 2WikiMultiHopQA test sets in Figure~\ref{fig:DatasetHops}. The statistics show that there is a high proportion of complex reasoning queries with 3 hops or more (aboout 30\%, 50\%, 25\%). HotpotQA lacks explicit hop annotations, so we instead count the number of supporting facts. MuSiQue has a significantly higher proportion of 3-hop and 4-hop queries compared to the other datasets, indicating great reasoning complexity. This observation is further corroborated by our experimental results in Table \ref{tab:main-result} and Figure \ref{fig:scaling-datasets}. The performance of our approach on MuSiQue is much lower than those of the other two datasets.

In the retrieval process, we employ the \textit{multilingual-e5-base}~\citep{wang2024multilinguale5} as the retriever and use the widely used Wikipedia dump from December 2018 as the retrieval corpus~\citep{karpukhin-etal-2020-dense} which comprises over 21 million passages. For generation, the default sampling parameters top-$p$, top-$k$ and temperature are set to 0.8, 50 and 0.7 respectively.
Evaluation metrics include Exact Match (EM), F1 score (F1), and Accuracy (Acc), where accuracy indicates whether the ground truth is a substring of the final generated answer. For reward model training, we sample 8,000 question-answer pairs from each dataset and generate more than 156,000 reasoning paths using our proposed AirRAG (rollouts=32, $n$=4, $q_{div}$=1.0). In inference scaling experiments, we sample maximum computation budgets $L_{\max}$ (e.g., 8k, 16k, 32k, 64k and 128k tokens). The $L_{\max}$  (maximum effective context length) denotes the maximum number of input tokens across all rollouts following ~\citep{yue2024iterdrag}. The predetermined maximum tree depth $d$ is set to 10, specifically indicating that the SAY and SA actions are executed once, while the RA-QT or QT-RA actions have a maximum of 4 iterations.

To further substantiate our experimental approach, we present comprehensive details and rationales for the experimental design, including the parameter configurations for the rollout process in MCTS, the selection of the UCT function, reward computation strategies, and the implementation of baseline methods.

\noindent\textbf{Rollout Setting for MCTS}. In some experimental configurations, we adopt a rollout setting of $1$ for MCTS. A rollout value of $1$ does not imply that MCTS is not executed, but rather that only one complete simulation is performed at each decision step to estimate the value of the current node. This setting reduces computational cost and allows for rapid generation of preliminary results, which is especially beneficial in resource-constrained scenarios. Figure~\ref{fig:DocSize_Rollout_Scaling} presents the performance across rollout values ranging from $1$ to $32$, while Table~\ref{tab:main-result-n} reports results obtained with a rollout value of $32$.

\noindent\textbf{Contribution of MCTS}.
MCTS primarily operates on actions such as QT, RA, and SA, as well as parameters (e.g., $n$, $\mathrm{top}_p$, $\mathrm{top}_k$) within each action. As illustrated in Figure~\ref{fig:DocSize_Rollout_Scaling} and Table~\ref{tab:main-result-n}, applying MCTS with higher rollout values significantly improves performance compared to using a rollout value of $1$, and outperforms baselines such as IterDRAG and Search-o1.

\noindent\textbf{UCT and Implicit Priors}. We explore both UCT and PUCT as potential search strategies for MCTS and ultimately adopt UCT in our final implementation, owing to its simplicity and consistent empirical performance. Although UCT does not explicitly incorporate prior probabilities as in PUCT, our framework implicitly introduces a form of policy prior through dependencies between actions, which constrain action transitions. This implicit prior serves a function analogous to the explicit $P(s,a)$ calculation in PUCT, guiding the search toward high-quality actions based on prior reasoning steps. The complete action execution process is detailed in Section~\ref{sec:fundamental-actions}.

\noindent\textbf{Reward Computation and Backpropagation}. The reward score is determined by whether the final correct answer is obtained. Specifically, after executing action $A_5$ (\textit{Summary-Answer}), a reward score is computed for each leaf node. In our experimental setup, we consider scenarios in which the self-consistency sampling count is set to $1$ or $3$, as summarized in Table~\ref{tab:main-result} with $n_{\text{all}} = 1/3$. When $n_{\text{all}}=1$, the $A_5$ action generates a single answer and the leaf node is assigned a reward $Q$ of 1. For $n_{\text{all}}=3$, we apply the clustering method (e.g., jcdScore) described in Section~\ref{select_verify} to calculate reward scores for the three answer nodes. This method assigns higher scores to answers with greater confidence, consistent with the voting strategies commonly employed in mathematical tasks. We then use these computed scores to backpropagate and update the scores for all nodes along the search path. In the formula $Q(s_i, a_i) = Q(s_i, a_i) + Q(s_d, a_d)$, $Q(s_d, a_d)$ denotes the reward value associated with the final answer.

\noindent\textbf{Self-Consistency Baselines}. For the Vanilla RAG method, the improvement brought by self-consistency voting is limited. In our main experiments, we set the rollout parameter to $1$ to ensure a fair comparison and to avoid potential confounding effects arising from applying self-consistency solely to the final results.

\noindent\textbf{Context Length Expansion for IterDRAG}. For IterDRAG, we follow the protocol outlined in the original paper, modifying three key aspects: the number of retrieved documents (ranging from $3$ up to $300$ maximum), the number of contextual examples (fixed at $5$ in our experiments), and the number of iterations (up to $5$). Further details regarding these inference scaling strategies are discussed in the work by Google DeepMind~\citep{yue2024iterdrag}.

\section{Additional Experiment Results}\label{sec:additional_exp}

\begin{figure*}[t]
\centering
\includegraphics[width=0.98\textwidth]{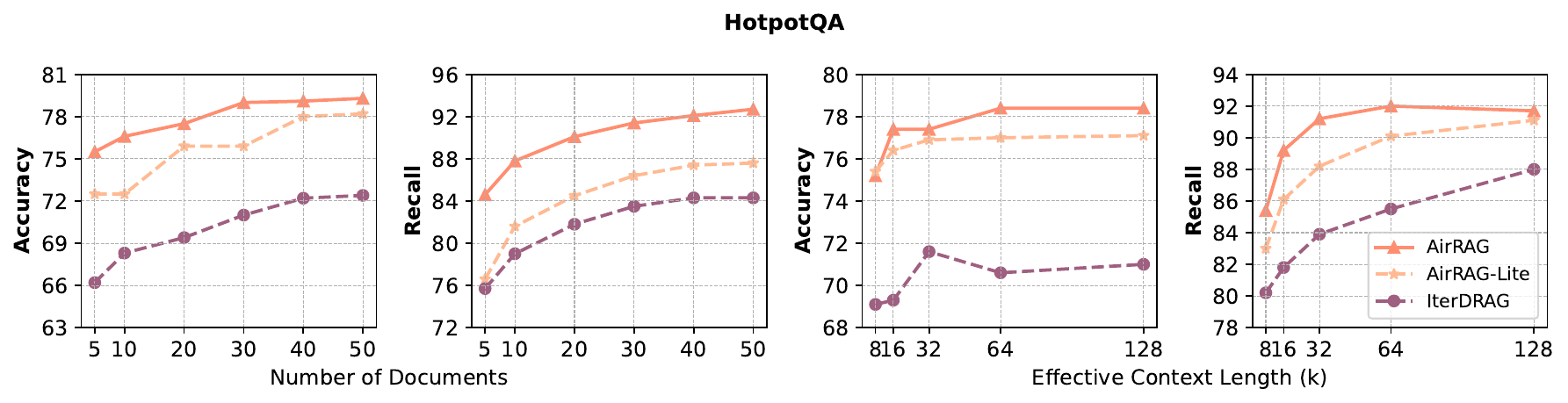}
\includegraphics[width=0.98\textwidth]{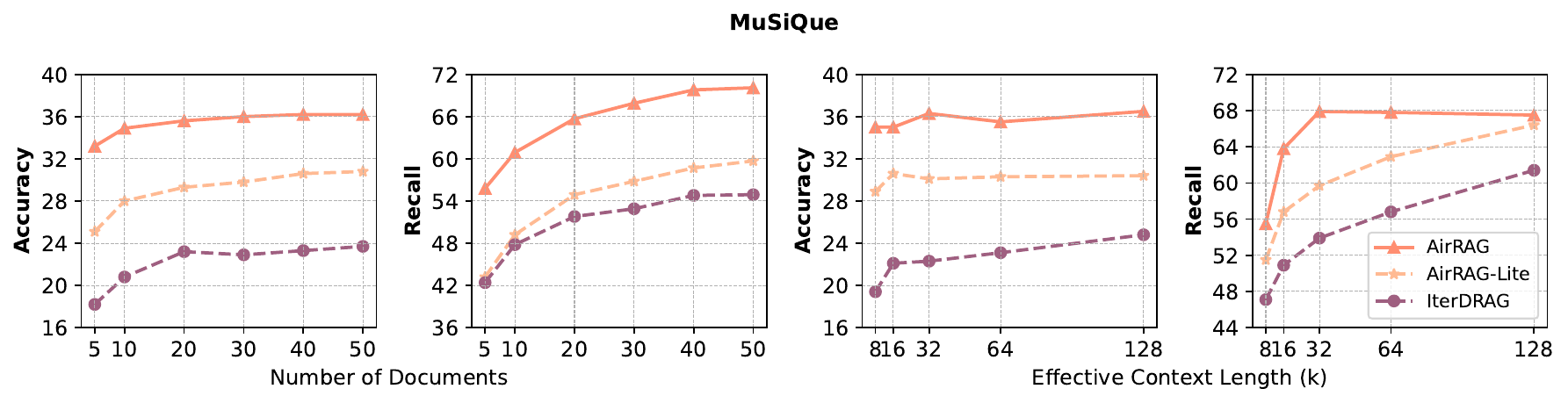}
\includegraphics[width=0.98\textwidth]{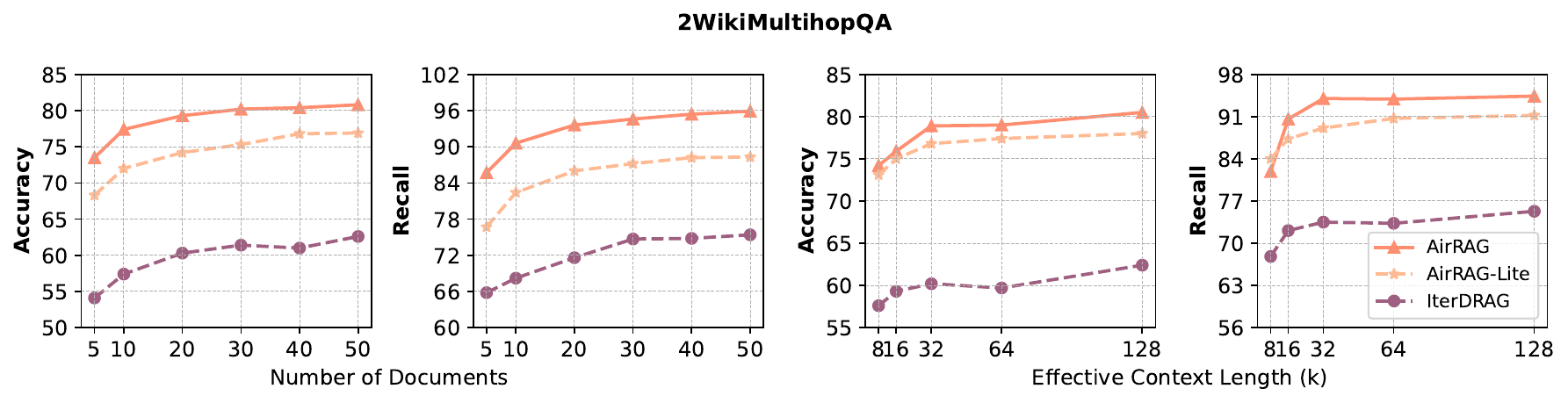}
\caption{Impact of the retrieved document number scaling and the maximum context length scaling over three datasets.}
\label{fig:scaling-datasets}
\end{figure*}

\begin{figure*}[htbp]
\centering
\includegraphics[width=0.98\textwidth]{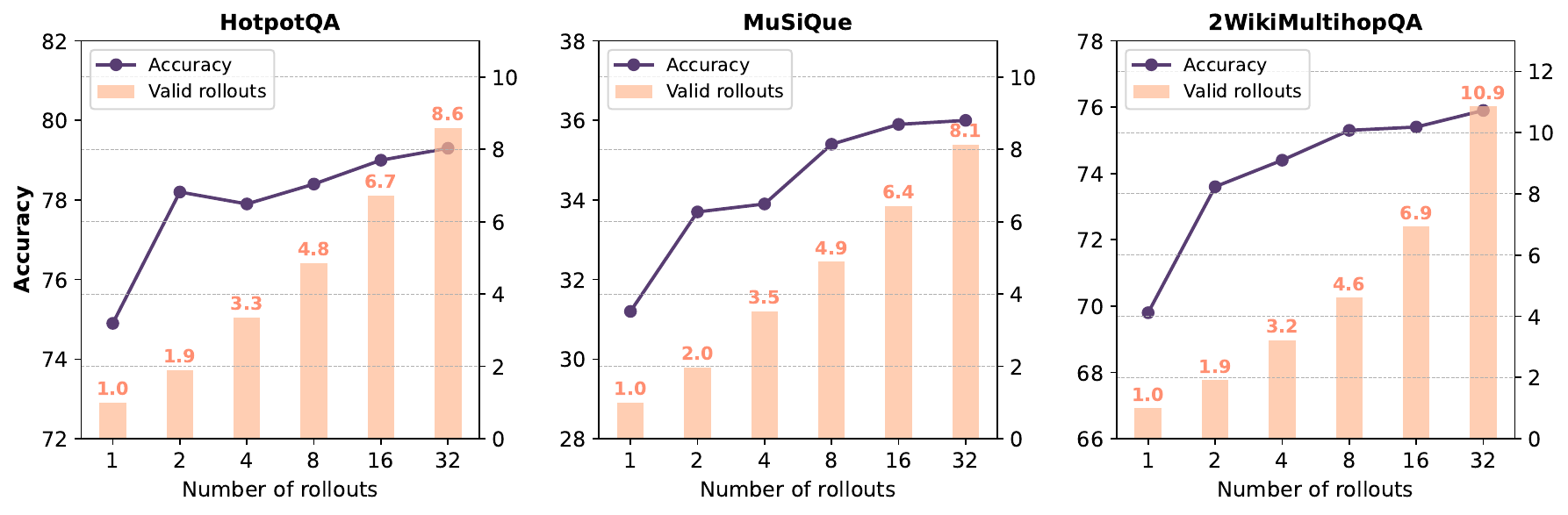}
\caption{Performance comparison on the number of different effective rollouts over three datasets. Sampling more diverse reasoning paths consistently improves accuracy.}
\label{fig:rollouts-datasets}
\end{figure*}

We evaluate the performance of AirRAG on various complex QA datasets. Table~\ref{tab:scaling-main-result} compares its accuracy and F1 with strong baselines under the given inference computation budget, which is implemented based on \textit{Qwen2.5-14B-Instruct} and one million document database. The optimal performance exhibits consistent gains as $L_{\max}$ expands, which is termed as the \textit{inference scaling laws} for RAG~\citep{yue2024iterdrag}. We integrate the remaining methods for a given maximum computational budget into our approach, dubbed as \textit{AirRAG-Blender}. The best results are obtained by using only the SA action to refine the final answer from all candidates, as shown in Table~\ref{tab:scaling-main-result}. This also demonstrates the flexibility of our approach architecture.
In addition, to verify the robustness and generalization of AirRAG, Table~\ref{tab:main-result-llama} shows the performance on more diverse LLMs and datasets. For a fair comparison, we utilize the widely used Wikipedia dump from December 2018~\citep{karpukhin-etal-2020-dense} as the retrieval corpus. We observe consistent improvements over vanilla RAG and existing iterative methods (more than 10\% on average). The significant boost over IterDRAG and Auto-RAG suggests that AirRAG explores more effective reasoning paths through the human-like thinking paradigm and tree-based search. Furthermore, we present detailed inference scaling results for each dataset individually, as shown in Figure \ref{fig:scaling-datasets} and Figure \ref{fig:rollouts-datasets}.

\begin{table*}[htbp]
 \centering
 \resizebox{0.78\textwidth}{!}{
\begin{tabular}{clcccccccc} 
\toprule
\multirow{2}{*}{$L_{\max}$} & \multirow{2}{*}{Method}  & \multicolumn{2}{c}{HotpotQA}  & \multicolumn{2}{c}{MuSiQue} & \multicolumn{2}{c}{2Wiki} & \multicolumn{2}{c}{\textbf{Average}} \\  
\cmidrule(l){3-10}  &  & F1 & Acc & F1 & Acc & F1 & Acc &\textbf{F1}& \textbf{Acc}\\ 
\midrule
\multirow{4}{*}{8k} 
    & \multicolumn{1}{l}{\text{ZeroShot QA}} & 42.5 & 41.3 & 13.5 & 12.1 & 48.2 & 47.3 & 34.7 & 33.6 \\
    & \multicolumn{1}{l}{\text{Vanilla RAG}} & 70.3 & 65.4 & 23.0 & 17.7 & 55.8 & 53.4 & 49.7 & 45.5 \\
    & \multicolumn{1}{l}{\text{IterDRAG}$^{*}$} & 74.3 & 69.1 & 26.7 & 19.4 & 60.5 & 57.6 & 53.8 & 48.7  \\
    & \multicolumn{1}{l}{\text{AirRAG-Lite}} & \textbf{80.6} & \textbf{75.4} & 35.4 & 28.9 & 75.3 & 73.1 & 63.8 & 59.1 \\
    & \multicolumn{1}{l}{\text{AirRAG}} & 79.6 & 75.2 & \textbf{41.0} & \textbf{35.0} & \textbf{76.0} & \textbf{74.2} & \textbf{65.6} & \textbf{61.5} \\
    \cmidrule(lr){2-10}
    & \multicolumn{1}{l}{\text{AirRAG-Blender}} & 81.1 & 79.8 & 41.6 & 36.4 & 82.2 & 81.7 & 68.3 & 66.0 \\
\midrule
\multirow{4}{*}{32k}    
    & \multicolumn{1}{l}{\text{Vanilla RAG}} & 77.1 & 72.0 & 29.0 & 22.9 & 60.9 & 58.1 & 55.7 & 51.0 \\
    & \multicolumn{1}{l}{\text{IterDRAG}$^{*}$} & 77.7 & 71.6 & 30.8 & 22.3 & 63.0 & 60.2 & 57.1 & 51.4  \\
    & \multicolumn{1}{l}{\text{AirRAG-Lite}} & 82.4 & 76.9 & 36.7 & 30.1 & 78.8 & 76.8 & 66.0 & 61.3 \\
    & \multicolumn{1}{l}{\text{AirRAG}} & \textbf{82.5} & \textbf{77.4} & \textbf{43.2} & \textbf{36.3} & \textbf{80.4} & \textbf{78.9} & \textbf{68.7} & \textbf{64.2} \\
    \cmidrule(lr){2-10}
    & \multicolumn{1}{l}{\text{AirRAG-Blender}} & 82.9 & 80.6 & 43.3 & 37.6 & 83.4 & 83.0 & 69.9 & 67.1 \\
\midrule
\multirow{3}{*}{128k}   
    & \multicolumn{1}{l}{\text{IterDRAG}$^{*}$} & 76.8 & 71.0 & 31.7 & 24.8 & 65.5 & 62.4 & 58.0 & 52.7  \\
    & \multicolumn{1}{l}{\text{AirRAG-Lite}} & 82.5 & 77.1  & 35.7 & 30.4 & 78.3 & 76.0 & 65.5 & 62.2 \\
    & \multicolumn{1}{l}{\text{AirRAG}} & \textbf{83.3} & \textbf{78.0} & \textbf{43.5} & \textbf{36.5} & \textbf{82.3} & \textbf{80.5} & \textbf{69.7} & \textbf{65.0} \\
    \cmidrule(lr){2-10}
    & \multicolumn{1}{l}{\text{AirRAG-Blender}} & 83.7 & 81.4 & 43.9 & 38.5 & 84.4 & 84.2 & 70.6 & 68.0 \\
\bottomrule
\end{tabular}}
 \caption{Overall evaluation results under different computational resource budgets, where \textit{Qwen2.5-14B-Instruct} is used as the generator LLM. * indicates the results reproduced by us. $L_{\max}$ denotes the maximum number of input tokens across all rollouts. The best results for each $L_{\max}$ are in \textbf{bold}. The number of both rollouts and output sequences is set to 1 for our proposed AirRAG methods.}
 \label{tab:scaling-main-result}
\end{table*}


\begin{table*}[htbp]
 \centering
 \resizebox{0.78\textwidth}{!}{
\begin{tabular} {@{}cccccccccc @{}} 
\toprule
\multirow{2}{*}{Method}  & \text{NQ} & \text{TriviaQA} & \text{PopQA} & \text{WebQA} & \text{HotpotQA}  & \text{2Wiki}\\  
\cmidrule(l){2-7}  & \text{EM} & \text{EM} & \text{F1} & \text{EM} & \text{F1} & \text{F1}\\ 
\midrule
\multicolumn{1}{l}{\text{Vanilla RAG}} & 35.1 & 58.8 & 36.7 & 15.7 & 35.3 & 21.0\\
\multicolumn{1}{l}{\text{Self-RAG}} & 36.4 & 38.2 & 32.7 & 21.9 & 29.6 & 25.1\\
\multicolumn{1}{l}{\text{Iter-RetGen}} & 36.8 & 60.1 & 37.9 & 18.2 & 38.3 & 21.6  \\
\multicolumn{1}{l}{\text{Auto-RAG}} & 37.9 & 60.9 & 47.8 & 25.1 & 44.9 & 48.9  \\
\multicolumn{1}{l}{\text{AirRAG}} & \textbf{53.6} & \textbf{63.2} & \textbf{51.8} & \textbf{52.6} & \textbf{67.6} & \textbf{66.3}  \\
\bottomrule
\end{tabular}
}
 \caption{Performance comparison on six benchmarks, where \textit{Llama3-8B-Instruct} is used as the generator LLM. Partial experimental results are quoted from \citet{FlashRAG} and \citet{yu2024autorag}. The best results are in \textbf{bold}. The number of both rollouts and output sequences is set to 1. The number of documents for a single retrieval is set to 5. Our proposed AirRAG significantly outperform the others.}
 \label{tab:main-result-llama}
\end{table*}


\begin{figure*}[t]
\centering
\includegraphics[width=0.6\textwidth]{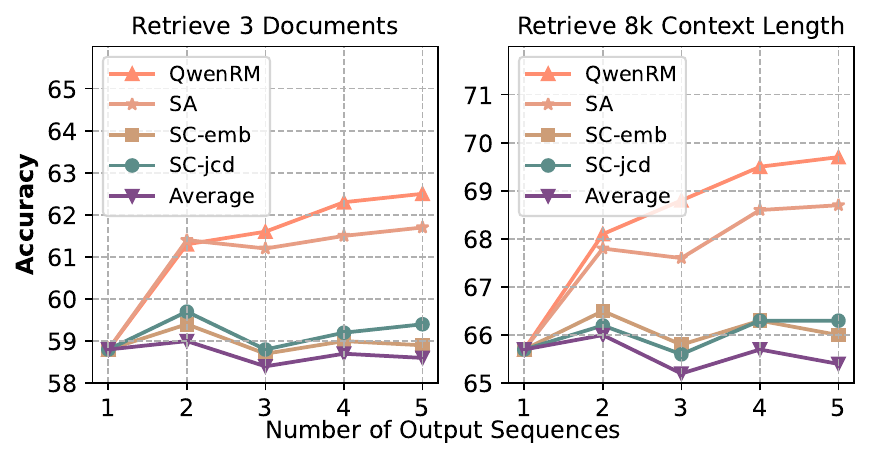}
\caption{Performance comparison of different verification methods. "QwenRM" is short for reward model trained on the Qwen model. "SA" is the reasoning action of summary and answer. "SC-emb/jcd" are two self-consistency verification methods based on text embeddings and jaccard similarity. "Average" is the average score over all candidate rollouts. The single retrieval process is set to retrieve three documents or fixed 8k context.}
\label{fig:verify}
\end{figure*}

\section{Prompt Examples}\label{sec:prompt_examples}
Given a user input query, our proposed AirRAG, as shown in Figure~\ref{fig:framework}, first attempts the \textit{direct answer} (DA) action without prompts and performs \textit{system analysis} (SAY) using the prompt in Figure~\ref{fig:say_prompt}. Subsequently, AirRAG performs \textit{retrieval and answer} (RA) with the prompt in Figure~\ref{fig:ra_prompt}, or \textit{query transformation} (QT) to generate refined queries for better retrieval and answer. This process of RA-QT or QT-RA can continuously iterate until no new sub-queries arise or the maximum iteration depth is reached. Finally, the \textit{summary answer} (SA) in Figure~\ref{fig:ra_prompt} utilizes all the information and conclusions from intermediate steps to refine the final answer.

\section{Case Study}\label{sec:case_study}
To further illustrate our approach, we select a representative sample from the complex multi-hop dataset MuSiQue for detailed analysis, as shown in Figures~\ref{fig:iterative_ambiguous_query} and~\ref{fig:air_rag_case}. Figure~\ref{fig:air_rag_case} specifically visualizes a reasoning path generated by our method. The SAY action can produce various expressions, and the QT action may generate multiple query formulations, leading to a diverse set of reasoning nodes and paths. This diversity enables our model to retrieve key information fragments that support both intermediate steps and the final answer. Compared with single-path generation under a fixed retrieval environment, such multi-path exploration substantially increases the likelihood of identifying crucial supporting evidence.

\begin{figure*}[htbp]
\begin{tcolorbox}[
    colback=mywhite,
    colframe=myblue,
    title=Example prompt for SAY reasoning action]
Given the user query, you may rephrase it for better clarity, summarize it at a higher level, or decompose it into multiple sub-queries to facilitate more effective information retrieval and response generation. If no modification is necessary, return "None". Otherwise, list sub-queries, each on a new line.\\
<Here are some examples.>\\
Query: \{question\}\\
Output: 
\end{tcolorbox}
\caption{Example prompt for SAY reasoning action.}
\label{fig:say_prompt}
\end{figure*}

\begin{figure*}[htbp]
\begin{tcolorbox}[
    colback=mywhite,
    colframe=myblue,
    title=Example prompt for QT reasoning action]
Given the context provided, please determine whether rephrasing, summarization, or decomposition into sub-queries is necessary to enhance the accuracy and efficiency of information retrieval and response generation. If no modification is required, return "None". Subsequent queries should be listed individually.\\
<Here are some examples.>\\
Main Query: \{question\}\\
History: \{history\}\\
This Query: \{this\_question\}
\end{tcolorbox}
\caption{Example prompt for QT reasoning action.}
\label{fig:qt_prompt}
\end{figure*}

\begin{figure*}[htbp]
\begin{tcolorbox}[
    colback=mywhite,
    colframe=myblue,
    title=Example Prompt for RA actions]
You are an expert in question answering. I am going to give you some contexts with may or may not be relevant to the question. Answer the question according to the contexts.\\
\{contexts\}\\
Question: \{question\}
\end{tcolorbox}
\caption{Example Prompt for RA actions.}
\label{fig:ra_prompt}
\end{figure*}

\begin{figure*}[ht]
\begin{tcolorbox}[
    colback=mywhite,
    colframe=myblue,
    title=Example prompt for SA reasoning action]
You are an expert in question answering. Given the context, sub-queries and responses, output a correct and concise answer to User Query.\\
<Here are some examples.>\\
User Query: \{question\}\\
\{history\}\\
Contexts: \{contexts\}\\
Final Answer: 
\end{tcolorbox}
\caption{Example prompt for SA reasoning action.}
\label{fig:sa_prompt}
\end{figure*}


\begin{figure*}[htbp]
\begin{tcolorbox}[
    colback=gray!20,  
    colframe=black,
    title= The iterative methods often traps in a solution space
    ]
\small
Context: <Retrieved documents>\\
Question: What is the mouth of watercourse for the body of water where Bartram's Covered Bridge is located?\\
\textbf{Follow up: Which body of water is Bartram's Covered Bridge located on?}\\
Follow up: | Intermediate answer: | So the final answer is:\\

\textbf{Output:} Follow up: Which body of water is Bartram's Covered Bridge located on?\\

(Note: The context does not specify which body of water Bartram's Covered Bridge is located on, but it does mention that the bridge carries Goshen Road over Crum Creek. Therefore, we \textbf{need to confirm that Crum Creek is the body of water in question before proceeding to find the mouth of the watercourse.})\\

$\cdots$\\
$\cdots$ (It enters an endless cycle of iteration until it reaches predefined maximum depth.)
\tcblower
\textbf{Gold Answer:} Delaware River\\
\textbf{Process Annotation:} Q1: Bartram's Covered Bridge >> crosses\\
A1: Crum Creek\\
Q2: Q1 >> mouth of the watercourse\\
A2: Delaware River\\

The iterative self-exploration often traps in a solution space with low-quality reasoning steps even after many attempts. 
\end{tcolorbox}
\caption{Incorrect reasoning path of iterative RAG methods. Ambiguity or missing information in the intermediate steps leads in duplicate sub-query.}
\label{fig:iterative_ambiguous_query}
\end{figure*}

\begin{figure*}[htbp]
\begin{tcolorbox}[
    colback=gray!20,  
    colframe=black,
    title=The simple and effective action space of AirRAG leads to the correct reasoning trajectory,
    ]
\small
Question: What is the mouth of watercourse for the body of water where Bartram's Covered Bridge is located?\\

\textbf{Action1: SAY}\\
Input: What is the mouth of watercourse for the body of water where Bartram's Covered Bridge is located?\\
Output: \textbf{Where is Bartram's Covered Bridge located?\\
What is the mouth of the watercourse for the body of water near Bartram's Covered Bridge?}\\

\textbf{Action2: RA}\\
Input: <Retrieved documents> \quad Where is Bartram's Covered Bridge located?\\
Output: Bartram's Covered Bridge is located on the border between Delaware County and Chester County, Pennsylvania, carrying Goshen Road over Crum Creek.\\

\textbf{Action3: QT}\\
Input: What is the mouth of the watercourse for the body of water near Bartram's Covered Bridge?\\
Output: \textbf{What is the mouth of Crum Creek?}\\

\textbf{Action4: RA}\\
Input: <Retrieved documents> \quad What is the mouth of Crum Creek?\\
Output: Based on the provided contexts, there are two different locations referred to as Crum Creek, and they flow into different bodies of water:\\
1. The Crum Creek in Delaware County and Chester County, Pennsylvania, flows into the Delaware River in Eddystone, Pennsylvania.\\
2. The Crum Creek (also called Cathatachua Creek) in East Creek, New York, flows into the Mohawk River.\\
Since the question does not specify which Crum Creek is being referred to, both answers are possible. However, if we consider the first context as the primary reference, the mouth of Crum Creek is the Delaware River in Eddystone, Pennsylvania.\\

\textbf{Action5: SA}\\
Input: <Retrieved documents> \quad <Sub-queries with answers>\\
Output: The mouth of the watercourse for the body of water where Bartram's Covered Bridge is located, \textbf{Crum Creek}, is the \textbf{Delaware River} in Eddystone, Pennsylvania.
\tcblower
\textbf{Gold Answer:} Delaware River. \\

AirRAG obtains the correct answer by performing five reasoning actions efficiently.
\end{tcolorbox}
\caption{Effective reasoning rollout of AirRAG. The fundamental action space and tree-based search provides greater compatibility and flexibility, which avoids being trapped in a single solution space by a low-quality intermediate step for chain-like reasoning paradigm.}
\label{fig:air_rag_case}
\end{figure*}

\end{document}